\documentclass[10pt,twocolumn,letterpaper]{article}
\usepackage[]{iccv}
\usepackage{helvet}  
\usepackage{courier}  
\usepackage[hyphens]{url}  
\usepackage{graphicx} 
\usepackage{times}
\usepackage{epsfig}
\usepackage{graphicx}
\usepackage{caption}
\usepackage{pifont}
\usepackage{tabularx}
\usepackage{enumitem}
\usepackage{makecell}

\setlist[itemize]{noitemsep, topsep=0pt}

\usepackage[pagebackref=true,breaklinks=true,letterpaper=true,colorlinks,bookmarks=false]{hyperref}

\iccvfinalcopy 

\def\iccvPaperID{2186} 

\usepackage{newfloat}
\usepackage{listings}
\DeclareCaptionStyle{ruled}{labelfont=normalfont,labelsep=colon,strut=off} 
\usepackage{comment}
\usepackage{amsmath,amssymb} 
\usepackage{color}
\usepackage{booktabs} 
\usepackage{xcolor}

\ificcvfinal\fi

\usepackage{color,soul}

\usepackage[linesnumbered,ruled,vlined]{utils/algorithm2e}
\include{utils/macros}

\title{Building Scalable Video Understanding Benchmarks through Sports} %

\author{
Aniket Agarwal\textsuperscript{\rm 1}\thanks{Equal Contribution} \thanks{Done while interning at Princeton and University of Toronto}\quad
Alex Zhang\textsuperscript{\rm 2}\footnotemark[1]\quad
Karthik Narasimhan\textsuperscript{\rm 2}\quad
Igor Gilitschenski\textsuperscript{\rm 3}\quad
\\
Vishvak Murahari\textsuperscript{\rm 2}\thanks{Equal Advising}\quad
Yash Kant\textsuperscript{\rm 3}\footnotemark[3] 
 \vspace{3pt} \\
\textsuperscript{\rm 1}IIT Roorkee \quad
\textsuperscript{\rm 2} Princeton University \quad
\textsuperscript{\rm 3}University of Toronto \quad
 \vspace{3pt} \\
\textcolor{blue}{\url{https://asap-benchmark.github.io/}}
}

\begin{document}

\maketitle

\begin{abstract}


Existing benchmarks for evaluating long video understanding falls short on two critical aspects, either lacking in scale or quality of annotations. These limitations arise from the difficulty in collecting dense annotations for long videos, which often require manually labeling each frame. In this work, we introduce an automated Annotation and Video Stream Alignment Pipeline (abbreviated ASAP). 
We demonstrate the generality of ASAP by aligning unlabeled videos of four different sports with corresponding freely available dense web annotations (\ie commentary). 
We then leverage ASAP’s scalability to create \lcric, a large-scale long video understanding benchmark, with over 1000 hours of densely annotated long Cricket videos (with an average sample length of $\sim$50 mins) collected at virtually zero annotation cost.  We benchmark and analyze state-of-the-art video understanding models on \lcric through a large set of compositional multi-choice and regression queries. We establish a human baseline that indicates significant room for new research to explore. 
Our human studies indicate that ASAP can align videos and annotations with high fidelity, precision, and speed. The dataset along with the code for ASAP and baselines can be accessed here: \textcolor{blue}{\url{https://asap-benchmark.github.io/}}.


\end{abstract}

\vspace{-10pt}
\section{Introduction}

\begin{figure}
    \centering
    \includegraphics[width=0.45\textwidth]{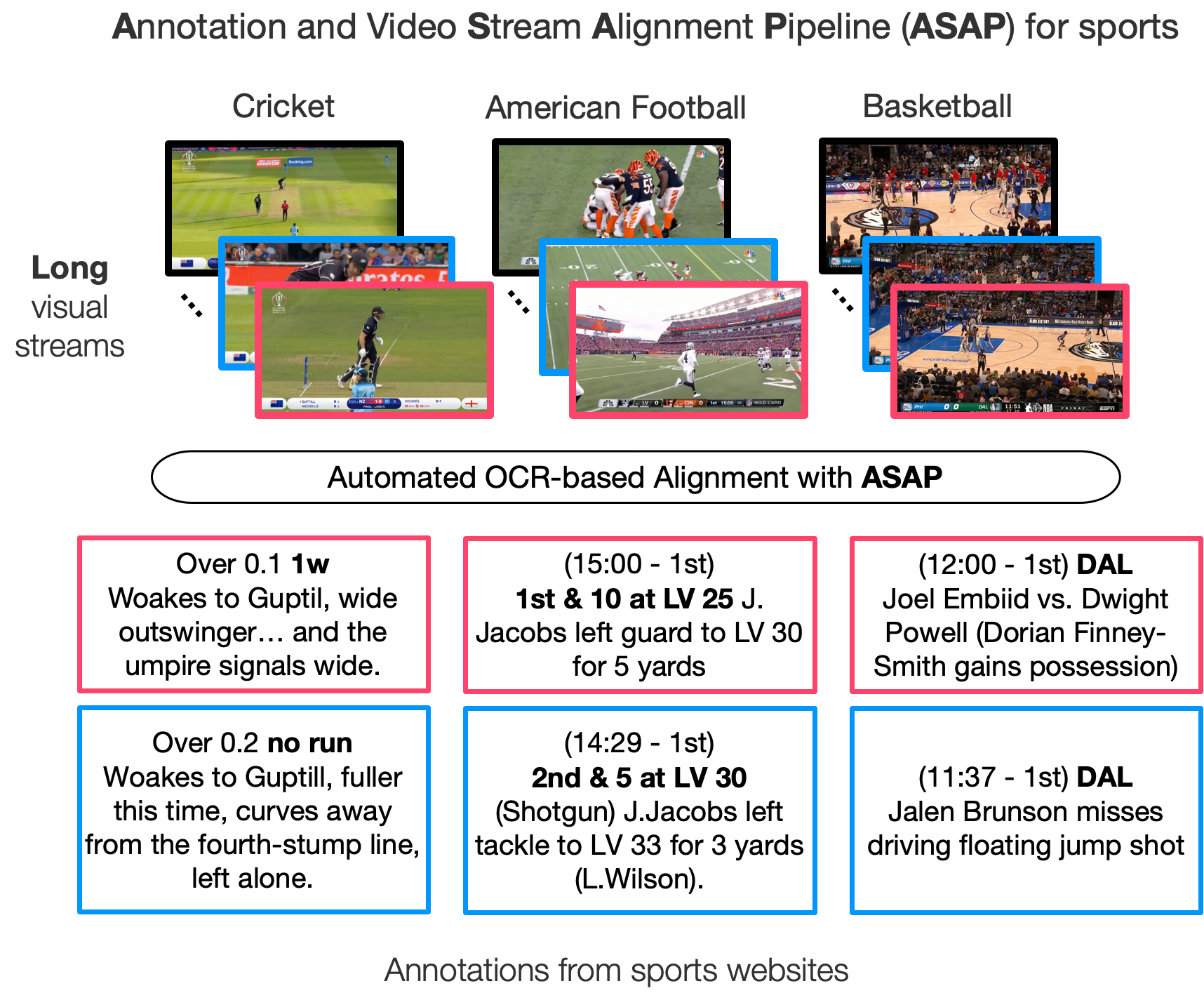}
    \captionof{figure}{ \textbf{Overview.} We propose the Annotation and Video Stream Alignment Pipeline (ASAP) for sports to align unlabelled sports videos with structured annotations publicly available on the web using an OCR based module. These newly aligned annotations can help compose structured queries that test long-horizon video understanding skills.}
    \label{fig:teaser}
\end{figure}


\noindent Humans learn and master skills (\eg playing guitar) by associating and reasoning over episodic memories captured over days, months, and years of failed and successful attempts~\cite{byrne2008learning}. Thus, building systems capable of understanding and reasoning over very long streams of visual data is a long-standing and crucial problem in Computer Vision. 

Long-horizon Video Understanding (\LVU) is the problem of reasoning over a long stream of video data, such as understanding the plot of a movie or analyzing the performance of a player in a lengthy game. Progress toward \LVU has been greatly limited by the lack of densely annotated data. Creating an \LVU benchmark requires manually annotating videos frame-by-frame, which is incredibly tedious and hard to scale. This constraint has limited the length of existing densely-annotated video understanding benchmarks (Table~\ref{tab:datasets}) from a few seconds~\cite{tgif, charades, vidsrl, xu2016msr} to a few minutes~\cite{zeng2016generation, krishna2017dense, wu_longvideo, ZhXuCoCVPR18, gella2018dataset, bain2020condensed}.

A line of previous works~\cite{MovieNet,MovieQA,lei2018tvqa} in \LVU have used readily available subtitles of TV shows or entire movies as dense annotations. While these videos are sufficiently long, manual annotations are still required to build non-trivial queries to evaluate \LVU skills~\cite{MovieQA,lei2018tvqa}, which greatly limits their scale. Another work~\cite{wu_longvideo} addresses this problem by extracting supervision from easily accessible YouTube metadata of nearly $\sim$30K movie clips spanning $1-3$ minutes. However, the annotated clips are short, and the proposed prediction tasks rely on noisy (and obscure) attributes (\eg  YouTube views, like-to-dislike ratio, and so on). 

Sports matches are a rich source of long videos (\eg a one-day Cricket match lasts nearly 8 hours) and usually have a brief scorecard embedded in the screen (shown in Figure \ref{fig:teaser}) that tracks the state of the match. Most sports matches also have dense annotations from experts available online (sports commentary describing major events in the game, \eg~\cite{ESPNCricinfo, ESPNfootball}). However, the annotations or the videos are not helpful individually unless they are aligned with each other.


Therefore, we introduce \ASAP, an automated annotation and video stream alignment pipeline, to automatically generate video datasets with dense annotations (i.e natural language commentary, major events in the match) by aligning arbitrarily long sports matches with their commentary freely available on the web~\cite{ESPNCricinfo, ESPNfootball}). \ASAP automatically parses \metadata from the scorecard embedded in sports match videos using an OCR detector \cite{OCR} and then uses this to automatically align the videos with dense annotations available on the web. To demonstrate the generality of \ASAP, we align unlabeled videos of four distinct sports (Cricket, Football/Soccer, Basketball, and American Football) with their corresponding web annotations, with an average of $95\%$ of the annotations being aligned within $\pm1$ second of their occurrence in the video.

We then leverage \ASAP's scalability to create \lcric, a large-scale \LVU benchmark with 1008 hours of densely annotated Cricket videos at virtually zero annotation cost, by auto-labeling 131 cricket matches of average length ~7.5 hours, containing nearly 475 timestamp recordings (balls per match) on average. To our knowledge, \lcric is the first automatically labeled sports video dataset that contains play-by-play annotations that span entire matches. To comprehensively evaluate \LVU on \lcric, we automatically curate multiple-choice (binary and N-way) and regression queries through simple composition with boolean operations, which require varying lengths of context to answer. These queries are complex and require context aggregation ranging anywhere from 5 minutes to an hour of continuous playtime (video). In the past, such compositional query building has been leveraged in popular vision and language datasets (\eg CLEVR~\cite{johnson2017clevr}, GQA~\cite{hudson2019gqa}).

We benchmark two recent state-of-the-art \LVU models TQN~\cite{zhang21}, MemViT~\cite{memvit} on \lcric, and find that their performance is significantly worse than human baseline ($\sim$38\% drop on query reasoning accuracy when evaluated on long clips containing $\sim$50 minutes of playtime). This demonstrates significant room for new research to explore. In summary, we make the following contributions:
\begin{itemize}
\item We propose \ASAP, an automated and scalable video labeling pipeline for aligning videos of sports matches of four different sports (Cricket, Football, Basketball, and American Football) with dense web annotations.
\item Using \ASAP, we create \lcric, a large-scale \LVU benchmark with 1008 hours of densely annotated Cricket videos with virtually zero annotation cost.
\item We benchmark the performance of two recent video understanding models on our dataset, provide ablations, and establish a human baseline on \lcric{}.
\end{itemize}

Finally, we note that the size of \lcric{} is only limited by access to videos of Cricket matches, and we foresee the dataset being much larger given access to more videos.

\begin{table}[t]
\centering
\resizebox{\linewidth}{!}{
\begin{tabularx}{0.65\textwidth}{l c c c c}
    \hline
    \textbf{Dataset} & \textbf{Avg clip} & \textbf{\# Annot.} & \textbf{\# Hours} & \textbf{Autolabel}\\
    \hline
    VidSitu ~\cite{vidsrl} & 10s & 145K & 81 & \ding{55} \\
    VideoStory ~\cite{gella2018dataset} & 18s & 123K & 396 & \ding{51} \\
    MSR-VTT ~\cite{xu2016msr} & 20s & \textbf{200K} & 41 & \ding{55} \\
    Charades  \cite{charades}  & 30s & 28K & 82 & \ding{55} \\
    TGIF \cite{tgif} & 30s & 126K & 86 & \ding{51} \\
    TVQA ~\cite{lei2018tvqa,lei2019tvqa} & 75s & \underline{152K} & 460 & \ding{55} \\
    VTW ~\cite{zeng2016generation}  & 90s & 45K & 213 & \ding{51} \\
    MovieClips ~\cite{bain2020condensed}  & 120s & 30K & \textbf{1270} & \ding{51} \\
    LVU ~\cite{wu_longvideo} & 120s & 11K & \textbf{1270} & \ding{51} \\
    YouCook II ~\cite{ZhXuCoCVPR18} & \underline{316s} & 15K & 176 & \ding{55} \\
    ActNet Captions~\cite{krishna2017dense} & 180s & 100K & 849 & \ding{55} \\
    \hline
    \textbf{LCric (ours)} & \textbf{2778s} & 62K & \underline{1008} & \ding{51} \\
\hline
\end{tabularx}
}
\caption{Comparison among annotated datasets for benchmarking video description and video understanding methods. \lcric has an average clip length of $\sim$2800 seconds, which is almost ten times larger than any previous work.}
\label{tab:datasets}
\end{table}


\section{Related Works}
%



%

\textbf{Existing benchmarks for LVU.} The paper~\cite{wu_longvideo} introduces the large-scale \LVU benchmark built on movie clips and metadata publicly available on YouTube. However, the videos only range from 1-3 minutes, and the annotations are limited due to their dependence on YouTube metadata. Other benchmarks for LVU include~\cite{virat}, which collected 29 hours of surveillance footage and bounding box annotations of major events but only have clips of length up to 3 minutes. Similarly,~\cite{meva} collected 144 hours of surveillance footage by hiring actors to enact predefined scripts but only has clips of length up to 5 minutes. The~\cite{ava_kinetics} benchmark collected 430 videos, each 15 minutes long, and dense bounding box information for 80 different atomic actions. Although their videos are relatively long, the annotated videos are only up to 15 minutes long, which is shorter than our annotated videos which are up to 45 minutes long, and are generated with no additional cost. Additionally,~\cite{LoL} collected the LoL dataset comprising 230 clips from the League of Legends video game, with each clip ranging from 30 to 50 minutes. However, they collected video highlight annotations based on very noisy and unreliable audience chat statistics. Video games also tend to have easy visual cues before major highlights that incentivize models to learn spurious correlations. In contrast, our tasks, by construction, force models to reason over a long horizon of events in a match.

\textbf{Collecting dense annotations for videos.} Annotating video datasets is extremely expensive. Prior works~\cite{xu2016msr, vidsrl, charades} have collected expensive annotations through Amazon Mechanical Turks (AMT) to label their clips with an associated text description, which greatly limits their scale (Table~\ref{tab:datasets}). Another line of work bootstraps from pre-existing annotations to generate new annotations. Other prior works bootstrap from pre-existing annotations to generate new annotations. For example, \cite{bain2020condensed} use existing captions on YouTube and IMDb metadata to label $30000$ movie clips, but assume these labels span the entirety of their clips. Similarly, \cite{zeng2016generation} take user-generated titles as labels for $18100$ user-generated clips, but again assume that these labels span the entirety of their clips. While \cite{gella2018dataset} temporally align sentences from paragraph captions to social media videos to form annotated clips. In contrast, our dataset is densely annotated by temporally aligning publicly available sports annotations, which offer more structure than text descriptions and are therefore hierarchically composable, enabling the creation of queries that require large but dense context. Although \cite{liang2010_context, liang2010, jwang2006} also align sports videos to online commentary information, they use heuristic methods that are not as accurate and do not scale well for generating longer and more video matches.



\textbf{Video datasets based on sports.} \noindent Recent interest in using computer vision to drive sports analytics~\cite{deepmind_football} suggests the importance of a dense annotation pipeline for sports videos. Current methods for producing sports datasets involve some form of manual annotations like in~\cite{sports-videos-in-the-wild-svw-a-video-dataset-for-sports-analysis}, where they manually label $4100$ sports clips based on the given action. Several works~\cite{openttgames, posetrack, kazemi2012using} have used automatically generated, densely labeled pose annotations for sports videos but are not easily scalable because they run computationally expensive, frame-level models to generate their annotations. Larger datasets such as~\cite{ucf101, sports-1m} exist but primarily focus on action recognition over a single clip, rather than a full sports video. Our dataset focuses on producing play-by-play annotations spanning entire sports match. Our general annotation pipeline can also be easily extended to the creation of video datasets for other sports.




\textbf{Video understanding models.} \noindent Processing long videos is challenging, as it requires aggregating context over long horizons with limited computational and memory budgets. In SlowFast networks~\cite{slowfast}, they use a dual pathway operating at a low and high frame rate to enable the aggregation of context over longer horizons while capturing low-level visual attributes. Meanwhile,~\cite{x3d} introduce a simple technique for progressive architecture expansion (along axes such as temporal, depth, width, etc.), inspired by feature selection in machine learning to achieve efficient models. Taking advantage of the implicit nature of transformers to handle long-range data~\cite{timesformer} proposes to adapt the standard transformer architecture for videos by enabling spatiotemporal feature learning directly from a sequence of frame-level patches. In MeMViT~\cite{memvit}, they introduce a memory-augmented multi-scale vision Transformer, greatly improves temporal support with minimal memory overhead, and achieves state-of-the-art performance on a variety of video understanding benchmarks. While the trend shows the model's capacity to handle longer and longer video clips more efficiently, the absence of a truly long-horizon dataset inhibits a fair comparison between these baselines and also inhibits the model's transferability to real-world video understanding tasks.

\textbf{Automated annotation pipelines.} \noindent Automating annotation pipelines, even partially, is critical to developing large-scale datasets. For example, the SBU dataset~\cite{sbu_neurips2011} for image-text retrieval pruned and paired Flickr queries with a set of images, the Conceptual Captions dataset~\cite{conceptual_captions} for image captioning leveraged the ``Alt-text" HTML attribute in web images, and the RedCaps dataset~\cite{redcaps}  harvested $12$ million image-text pairs from curated subreddits.
Meanwhile,~\cite{PontTuset_eccv2020} partially automate their annotation pipeline and collect multi-modal image annotations by asking annotators to describe an image through audio while simultaneously hovering their mouse over the region they are describing. We hope that our fully automated pipeline, \ASAP, will help create long and densely annotated video datasets at an unprecedented scale.

\begin{figure}[t!]
    \centering
    \includegraphics[width=\columnwidth]{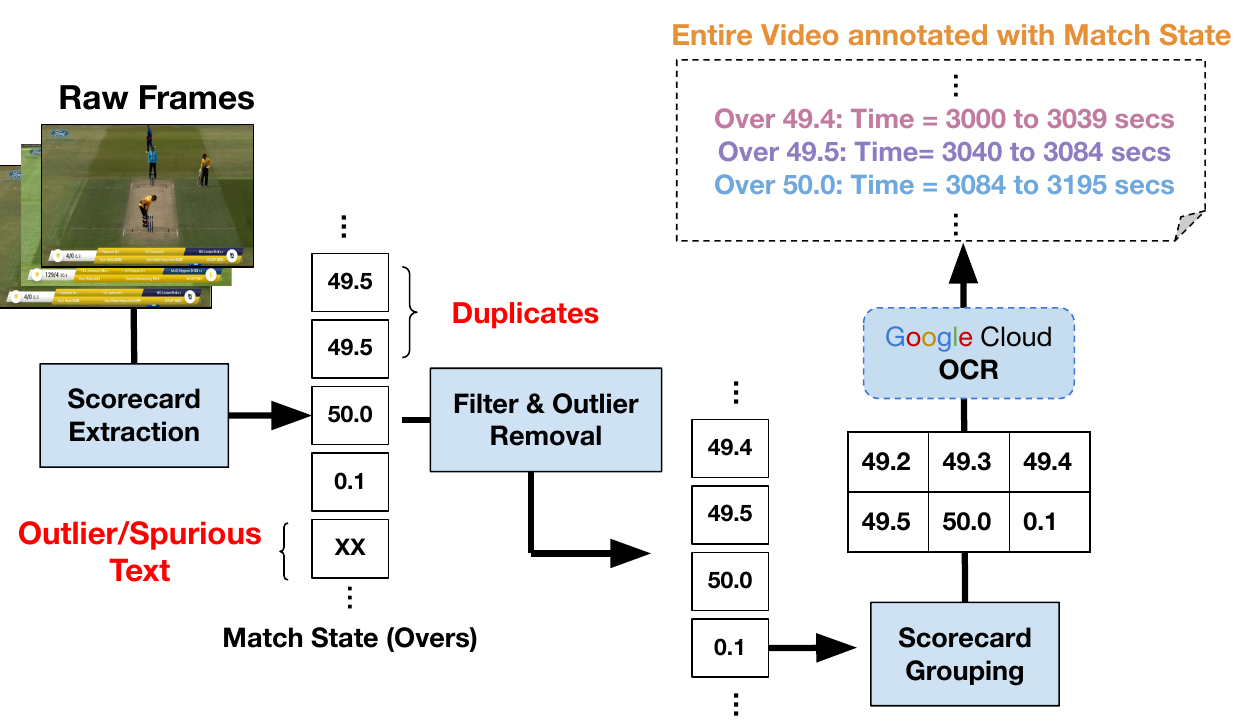}
    \caption{\textbf{Match State Extraction in Cricket.} \textbf{Left:} To extract the match state we begin by detecting scorecard bounding box across all video frames, and cropping its content out. \textbf{Middle:} We filter duplicates and outliers from extracted scorecards using L1 distance metric. \textbf{Right:} We batch the filtered patches in one image for the OCR call, and generate \textit{match states} aligned in time with the full video.}
    \label{fig:schematic}
\end{figure}

\begin{figure*}[t!]
    \centering
    \includegraphics[width=0.89\linewidth]{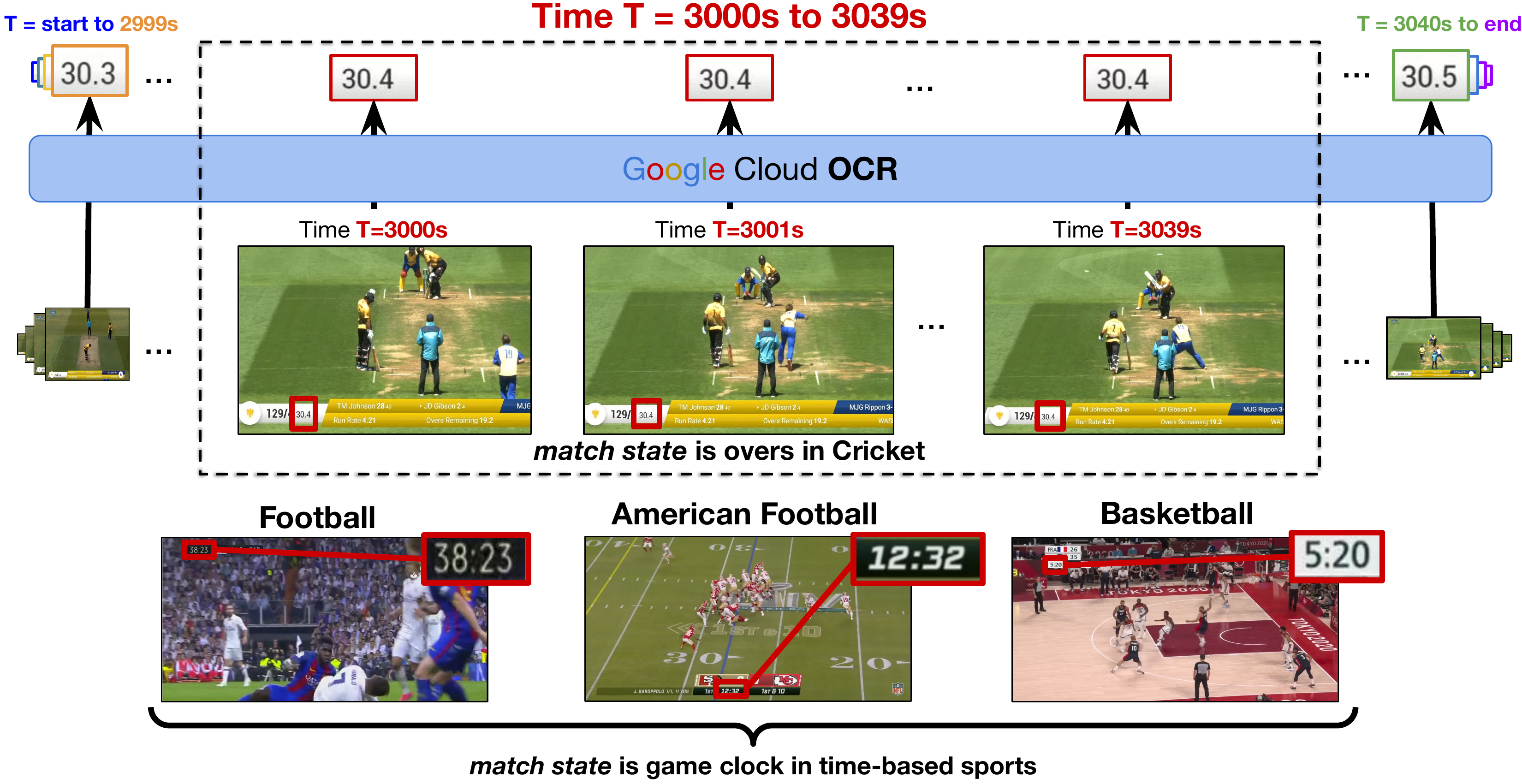}
    \caption{\textbf{Match States used by \ASAP for different sports.} \ASAP finds and stores the sequence of frames corresponding to each possible \metadata, which is necessary to temporally align web annotation data to its exact occurrence in each video. The process is as follows: \textbf{a) Top Row:} For each frame in a Cricket match video, \ASAP locates and extracts the \metadata (30.4 in the above example) from the scorecard of each frame using an \texttt{OCR} module. Next, \ASAP determines the sequence of frames (for \metadata= 30.4 in the example above, this would be the frames from T=3000 seconds to T=3039 seconds) that each \metadata corresponds to. \textbf{b) {Bottom Row:}} For other sports like American Football, Football, and Basketball, \ASAP uses the onscreen timer as \metadata. More generally, any temporal marker can be used as \metadata. }
    \label{fig:asap_stage1}
\end{figure*}




\section{\ASAP: \textbf{A}nnotation and Video \textbf{S}tream \textbf{A}lignment \textbf{P}ipeline for Sports Matches}

\noindent Sports matches provide an abundant source of long videos, along with a rich source of corresponding play-by-play annotations (i.e. expert commentary of major events in the match) easily accessible on the web~\cite{ESPNCricinfo, ESPNfootball}. 
These play-by-play annotations are, however, not useful standalone as they are not aligned with the video of the match. We, therefore, introduce \ASAP, a fully automated annotation pipeline for automatically aligning sports videos with their corresponding play-by-play annotations available on the web. \ASAP automatically parses \metadata from the scorecard embedded in sports match video through an OCR detector (as seen in Figure \ref{fig:schematic}) and then uses these states to align the video with dense annotations available on the web. Thus, \ASAP enables the creation of long video datasets with unprecedented scale at virtually zero annotation cost. We first give a brief overview of \ASAP and then describe the different stages of \ASAP in more detail.

\noindent \textbf{\ASAP  Overview.} \ASAP first extracts the scorecards embedded in a sports video and automatically parses the state of the match. For instance, Basketball and Football games have a running game clock, whereas, Cricket matches have information about the \textit{ball} being played (Section~\ref{subsec:cricket_overview}). We crop out the scorecard from each video frame in the match and extract the \metadata contained within it using an Optical Character Recognition (OCR) system (shown in Figure~\ref{fig:asap_stage1}). We then use the \metadata to get the corresponding commentary from the web annotations (Sec.~\ref{subsec:asap_stage2}) and assign one of the discrete events for that sport (Refer to Appendix~\ref{appendix:events} for a list of events). This multi-stage approach allows us to align every video frame in the match to the corresponding commentary and discrete event.

\subsection{Stage 1: Match State Extraction}
\label{subsec:asap_stage1}

\textbf{Extracting Match State.} In a sports video containing N video frames denoted as $[f_1, \cdots, f_N]$, we start by detecting a bounding box that encapsulates the \metadata (scorecard) information. For this, we sample a few frames uniformly throughout the video and run \textit{Google OCR} on them. Next, we determine the bounding box where text changes gradually across frames (i.e the bounding box containing the scorecard). Once these bounds are detected, we crop the scorecard contained within them across all the frames. We show example cropped scorecards in Figure~\ref{fig:asap_stage1}.

\textbf{Match State in Cricket (Overs).} For Cricket, we represent the match state by the \textit{ball} that is currently being delivered. For example, in the top row of Figure~\ref{fig:asap_stage1}, we extract “30.4” at frame $f_t$, which reveals that 4$^\text{th}$ ball of 30$^\text{th}$ over is being played in this frame. Assuming this ball (event) lasted for $n$ frames, we detect that at frame $f_{t+n}$ the match state changes to “30.5”, which means the event that took place on the 4th ball of 30$^\text{th}$ over lasted across frames $[f_t, f_{t+n}]$. In this manner, we can label all frames of the video sequence with a corresponding match state, and also club the consecutive frames across which the same match state persists.

\textbf{Match State in American Football, Football, and Basketball (Timer).} For remaining sports, we represent the match state by the onscreen timer usually displaying minutes and seconds passed. For example, in the bottom row of Figure~\ref{fig:asap_stage1}, we show the time extracted across different sports. As we use videos with a high frame rate of 30 FPS, the same match state persists across multiple consecutive frames and we, therefore, club these frames together like in Cricket.

We find that locating the scorecard across frames, as well as extracting \metadata from it is non-trivial -- due to the noisy and dynamic nature of the scorecard. We list some key challenges we encountered (and resolved) below.

\noindent \textbf{Occlusion, and changing attributes.} We found that scorecards can get occluded by advertisements, move to a different position on the screen, or change their attributes (format, shape, color) during gameplay. To address issues, we use a \textit{reference scorecard}, which is a template image containing an unoccluded crop of the scorecard. We compare this reference image against all cropped scorecards in the video using L1 distance, and remove spurious that lie far away from the reference. These are the frames \textbf{1)} being occluded by advertisements or other miscellaneous patches, \textbf{2)} having varying attributes of the scorecard (format, shape, color, etc.). We put frames rejected by this filtering in Appendix~\ref{appendix:asap}.


\noindent \textbf{\texttt{OCR} calls for every frame in a video can be expensive.} Sports matches last for many hours and passing all frames of a 30 FPS video through the paid \textit{Google OCR} becomes prohibitively expensive. To reduce the costs many-fold, we stack multiple scorecards into a single image and get multiple annotations with a single \texttt{OCR} call. We also skip frames that contain the same \metadata as an older frame by detecting changes in the scorecard between consecutive frames using a $L1$ distance metric.

\begin{figure}[t!]
    \centering
    \includegraphics[width=0.9\columnwidth]{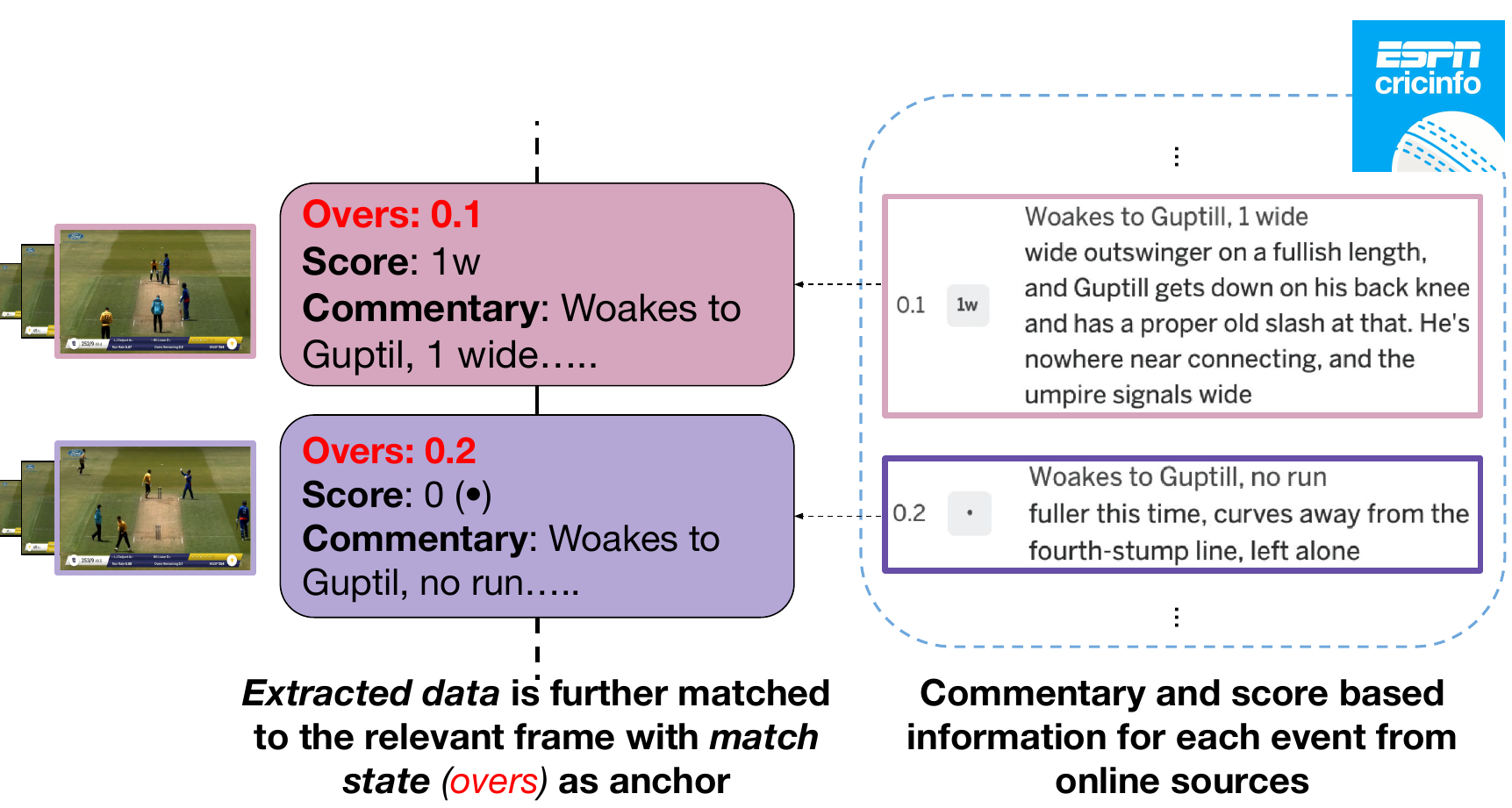}
    \caption{\textbf{\ASAP Stage 2 for Cricket matches.} \ASAP scrapes dense play-by-play annotations from the web and parses them into discrete events. \ASAP then uses the extracted \metadata from Stage 1 to align the annotations with the video of the match.}
    \label{fig:asap_stage2}
\end{figure}

\subsection{Stage 2: Aligning dense annotations with videos}
\label{subsec:asap_stage2}

Dense play-by-play annotations (i.e expert commentary, major events, etc.) are often easily available on the web~\cite{ESPNCricinfo, ESPNfootball}. Since these play-by-play annotations are indexed by the \metadata (precise \textit{play time} or \textit{ball} information), we map annotations based on their \metadata to their exact timestamps (frames) found in the first stage of \ASAP.

In addition to aligning the annotations with the video, ASAP also processes the sequence of play-by-play annotations into a sequence of discrete events, which we refer to as an \textit{event chain} (Figure~\ref{fig:asap_stage2}). While some sports (Cricket and Football) already contain discrete events (e.g. `foul', 'wicket', `boundary' etc.) in their annotations, for other sports (American Football and Basketball), we use string-matching to parse the commentary and assign each play to a fixed event that we define (e.g. `incomplete pass'). These extracted event chains can then be used as ground truth for evaluating \LVU models. Models can be queried on different segments of the event chain of varying lengths -- to test both short and long-horizon reasoning. We discuss the use of event chains for evaluation in Section~\ref{sec:query_formation}.

\subsection{\ASAP is faithful and fast}

\begin{figure}
    \centering
    \includegraphics[width=\columnwidth]{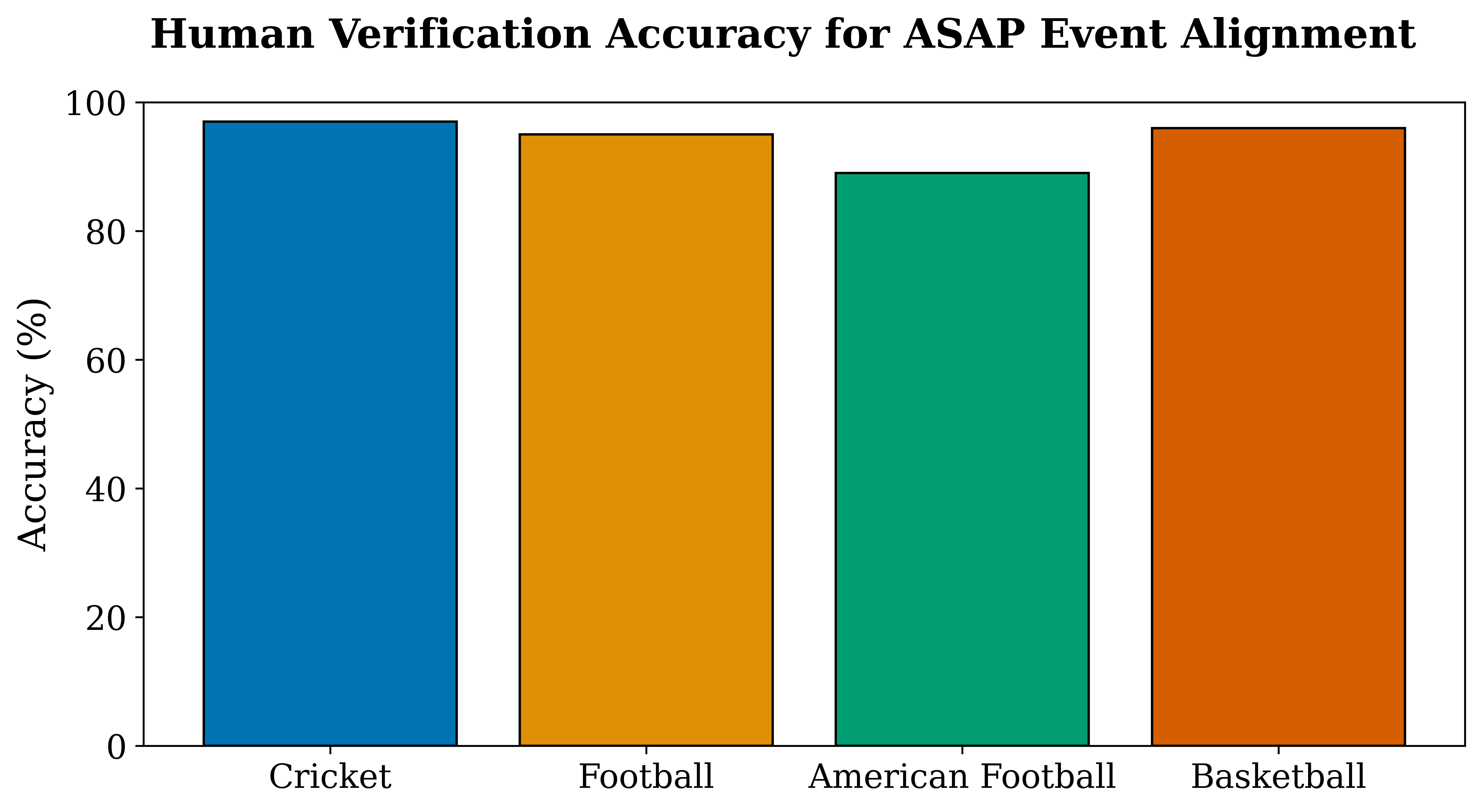}
    \caption{\textbf{Accuracy of ASAP.} Our human studies indicate \ASAP alignment to be highly accurate, with an average of $95\%$ of the annotations being correctly aligned to the corresponding video events ($\pm 1$ second) across 4 different sports.}
    \label{fig:amt_plot}
\end{figure}

\noindent{\textbf{How accurate is \ASAP?}} To verify \ASAP's ability to align dense annotations on the web with videos of sports matches, we conduct a study with human annotators on Amazon Mechanical Turk (AMT). We randomly sample clips from sports matches corresponding to 6 contiguous events in the event chain generated by \ASAP. For all the generated clips, we ask human annotators to provide timestamps for all 6 events and then check whether the provided timestamps belong to the intervals generated for those events by \ASAP. We plot the resulting accuracy of the timestamps in Figure \ref{fig:amt_plot} and find that \ASAP is highly accurate, with an average accuracy of 95.3\% across four very different sports, each differing in visual attributes, number of events, and length of plays. The drop in accuracy for American football annotations can be explained by the inconsistency in the timestamps provided by ESPN. For regular plays, the timestamp indicates when the play begins; however, for `touchdowns', the timestamp indicates when the team scored and not when the play begins. Additionally, penalties may affect the game clock, which we use to align our annotations, which sometimes leads to slight alignment issues for \ASAP. We present more details in Appendix~\ref{appendix:asap}.

\noindent{\textbf{How fast is \ASAP?}} \ASAP generates annotations with very high speed and requires just 10 minutes to align and process around 7 hours of video at 30FPS on a single machine. 

\section{Generating the \lcric dataset with \ASAP}

\noindent In this section, we describe how we leverage \ASAP to build a long video understanding (\LVU) dataset from Cricket videos. The \ASAP pipeline takes in videos of Cricket matches along with play-by-play commentary annotations from the web\footnote{\url{https://www.espncricinfo.com/}}. ASAP then aligns every video frame with the corresponding commentary annotation. Using the frame-aligned annotations produced by \ASAP, we describe a scalable approach for generating structured and compositional queries in Section \ref{sec:query_formation} to evaluate \LVU.  Our \lcric dataset is a collection of Cricket videos with play-by-play annotations and a set of auto-generated queries. 


\label{sec:lcric}

\subsection{Introduction to \cric}
\label{subsec:cricket_overview}
\noindent We present a brief introduction to the sport of \cric before discussing \lcric. \cric is played by two teams of 11 players each that alternate between \textit{batting} and \textit{fielding} throughout the game. The batting team aims to score \textit{runs} by hitting a ball bowled by the fielding team out of the playing field. Meanwhile, the fielding team aims to prevent the batting team from scoring runs and dismiss all players in the batting team by taking their \textit{wickets}. Each exchange where the fielding team bowls a valid ball and the batting team attempts to hit the ball to score runs is called a \textit{ball} (or \textit{delivery}) and a sequence of 6 \textit{balls} is called an \textit{over}. Each \textit{ball} is an \textit{atomic event} and there are 12 distinct possible events per ball, listed below:

\begin{itemize}
    \item The batting team scores $n$ runs, where $n \in \{0,...,9\}$
    \item A wicket is taken and the current batsman is dismissed.
    \item A wide (invalid) ball is bowled, giving the batting team an extra run and another ball.
\end{itemize}

We present a detailed discussion of different phases of Cricket in Appendix~\ref{appendix:lcric}.

\subsection{\lcric: Overview} 
\noindent Using \ASAP's scalability, we create \lcric, a large-scale \LVU benchmark with 1008 hours of densely annotated Cricket videos at virtually zero annotation cost, by auto-labeling 131 cricket matches of average length $\sim$7.5 hours, containing nearly 475 timestamp recordings (balls per match) on average. \ASAP automatically labels all the balls in a match with 1 of 12 events (Section~\ref{subsec:cricket_overview}) to generate a sequence of events (i.e \textit{event chain} shown in Figure~\ref{fig:asap_stage2}) for a cricket match. We then generate annotated video clips by segmenting the videos along with the aligned event chain into a contiguous sequence of 10-over ($\sim$50 minutes) clips.

\subsection{\lcric: Testing LVU via compositional queries} 
\label{sec:query_formation}

\noindent \textbf{Task Definition and \LVU Capabilities.} In an LVU task, the evaluated model is given a very long video clip, from a sports match of nearly 50 mins in our case, and it is tasked to answer a question (query) about it, e.g. “how many times did a wide ball occur in this video?”. An \LVU system needs to possess two types of skills: a) the ability to reliably detect local (short-term) events -- \eg, classifying an atomic event in Cricket (say wide, wicket, or run), and b) the ability to aggregate information across these local events given a task (which we refer to as a query) -- \eg, counting the total number of runs scored by the batting team in an arbitrarily long video. We evaluate these \LVU skills on \lcric, by automatically generating binary, multiple-choice, and regression queries, and evaluating them on long video segments.

\noindent \textbf{Min-Max occurrence query.} 
To test a model's ability to detect and remember events, we construct queries such as ``for a given video, did a wide ball (an event) occur between 3 and 5 times inclusive?". We generate these queries by sampling an atomic event from the set of all possible events, and then sampling two numbers, $o_{min}$ and $o_{max}$, to denote the minimum and the maximum number of occurrences needed for this query to be \textit{true}. 

\noindent \textbf{Binary queries by chaining occurrence queries.} To increase query diversity and complexity, we sample $n_{chain}$ different min-max occurrence queries and combine them using [and]/[or] operators. For example, for a given video spanning 10 overs ($\sim$50 mins), ``did a wide ball occur between 3 to 5 times [and] did a ball with 2 runs scored occur 1 to 3 times?". All \textbf{binary} queries in \lcric are formed by chaining 1-5 different min-max occurrence queries.

\noindent \textbf{Multiple-choice queries by counting occurrences.} We expand upon the binary occurrence queries by generating multiple-choice occurrence queries, which ask models to directly predict the number of occurrences rather than predicting membership in a range. An example of such a query is -- given a video, how many times did a wide ball occur after a ball with 4 runs? We note that these events are sequential, but not necessarily contiguous. As most non-trivial multiple-choice events in \lcric occur between 0-9 times in a given clip, we use $\{0,...,9\}$ as our answer choices.

\noindent \textbf{Filtering unbalanced queries.} We can compose many \LVU multiple-choice and binary queries using the above formulation, however, not all queries are balanced. Due to the rarity of certain events occurring in Cricket, some queries are far easier to guess correctly than others. For example, in a 45 minutes clip (spanning $\sim$ 10 overs), the query -- ``did a ball with 2 runs occur between 0 to 10 times" is true with a probability of $87\%$. We filter such queries based on the probability of their occurrence in training matches and ensure the average probability of occurrence of the selected queries to be between $0.45-0.55$ to avoid bias.

\noindent \textbf{Regression query for counting runs.} Lastly, we also experiment with a single regression query that asks the model to predict the number of runs scored as a regression output for a given video sequence.

\subsection{\lcric: Statistics and Dataset Splits}
\label{subsec:lcric_statistics}
\noindent{\textbf{Statistics.}} \lcric currently includes $1008$ hours of cricket match videos across $131$ unique matches (average length of 7.5 hours), along with $61957$ ball-by-ball annotations. All the videos are preprocessed at a resolution of 360p and we provide links to the source videos of higher resolution. 

\noindent{\textbf{Dataset splits}} To effectively test generalization, we split all the matches in \lcric into train, validation, and test splits and ensure a 3:1:1 ratio of the number of hours in each split. Due to a limited compute availability, we present ablations on a subset of \lcric and refer to this as \lcricm, which has around 420 hours of labeled Cricket matches, and therefore enables us to train ablations experiments in a shorter duration (2-3 days per experiment). We generate splits for \lcricm identically to \lcric. 

\section{Experiments on \lcric}

\noindent In this section, we describe our experiments benchmarking state-of-the-art \LVU models on \lcric. We also provide a human baseline on \lcric and demonstrate significant room for improvement. Finally, we analyze the performance of our baselines and present key insights to spur future modeling improvements on \lcric.

\subsection{Experimental Setup}

\noindent{\textbf{Preprocessing.}} We follow past work in \LVU~\cite{zhang2021temporal} by sampling videos at a lower FPS to make training over long videos feasible. We process our longest clips (containing 10 overs of the match) at 0.1 FPS, and process clips of 2-8 overs at 0.5 FPS respectively. We remove the scorecard from all frames to prevent annotation leakage and process frames at a resolution of 128 x 128.

\noindent{\textbf{Evaluation Metrics.}} We compute following evaluation metrics \textbf{1)} Classification accuracy for binary (\textbf{Binary Accuracy}) and multi-choice (\textbf{Multiple Accuracy}) queries \textbf{2)} Average L1 norm for regression queries (\textbf{Regression L1 Norm}).

\subsection{Baseline Models and Training Scheme}

\noindent Previous works \cite{fan2021multiscale, slowfast, x3d} in \LVU use pretrained CNNs~\cite{cnn} and Transformers~\cite{vaswani2017attention} paired with explicit memory modules for modeling long contexts. However, none of these methods can scale to video clips longer than a few minutes. Since our query set requires reasoning over contexts ranging up to an hour, we choose two state-of-the-art video understanding models, Temporal Query Network (TQN) \cite{zhang2021temporal} and MeMViT \cite{wu2022memvit} as benchmarks -- given their effective caching mechanisms for processing long videos. 

\noindent{\textbf{Temporal Query Network (TQN)~\cite{zhang2021temporal}}} TQN uses stochastic memory banks to efficiently model long horizon videos. They also introduce a transformer-based multi-query head to generate responses for multiple queries with a single pass through the network. TQN uses S3D~\cite{xie2018rethinking} for the visual backbone network and processes non-overlapping contiguous sequences of 8 frames with a temporal stride of 1. Please note that the term \textit{query} used in TQN is a modeling component and is not the same as our use of query in Section~\ref{sec:lcric}.

\noindent{\textbf{Memory-augmented Multiscale Vision Transformer (MeMViT)~\cite{wu2022memvit}}} MeMViT applies a memory caching strategy by processing videos in an online fashion, allowing the model to efficiently store context to reason over a long horizon. MeMViT builds upon ViT~\cite{vit} by using a novel pooling method and a dynamic patch resolution approach to reduce computational costs while processing long clips. We adapt MeMViT to handle our multi-query setting by leveraging the multi-query head introduced in TQN.

\noindent{\textbf{Training Schemes.}} We employ two different training schemes -- \textbf{1) Homogeneous training} where we train different models for the three different types of queries (binary, multi-choice, and regression) and \textbf{2) Mixed training} where we train a single model for all three types of queries. 

\subsection{Human Baseline}
\noindent To quantify the room for modeling improvements on \lcric, we measure the accuracy of human annotators through AMT~\cite{amt}. We provide annotators with video clips from \lcric and ask them to predict the sequence of ball-by-ball events (\ref{subsec:cricket_overview}). To compute human performance on our queries, we assume that given an event chain, humans can answer these queries by applying logical operators without mistakes. We detail our AMT setup in Appendix~\ref{ssec:amt}.



\subsection{Key Results}

\begin{figure}[]
    \centering
    \includegraphics[width=0.99\columnwidth]{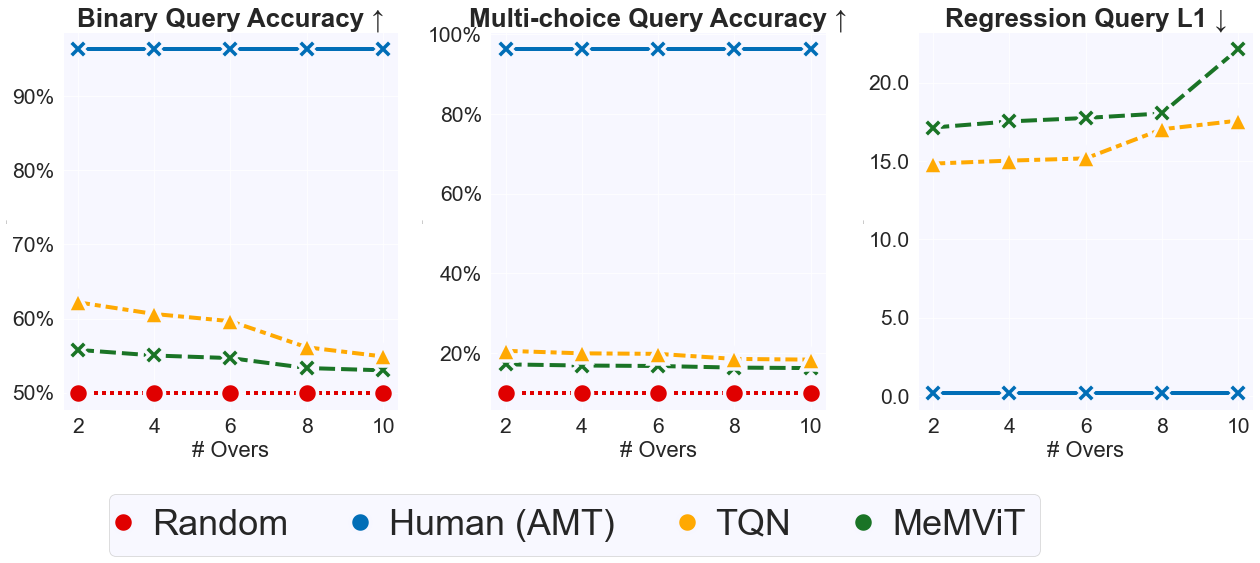}
    \caption{\textbf{Performance drops as video length (\# of overs) increase.} We plot the accuracy achieved by our baseline models for binary, multiple-choice, and regression queries across clips varying from 2-10 overs (length). \LVU models severely degrade in performance with longer time horizons and perform much worse than humans.}
    \label{fig:main_result}
\end{figure}

\noindent{\textbf{Performance of both TQN and MeMViT degrades rapidly and approaches random for very long clips}} To understand the impact of length of the videos on task performance, we train different baseline models for clips with over-lengths ranging from 2 overs ($\sim$10 minutes) to 10 overs ($\sim$50 minutes). Figure~\ref{fig:main_result} shows that performance rapidly decreases with increasing clip length and approaches the random baseline for binary and multi-choice queries. This result, in addition to the strong human baseline, shows significant room for modeling improvements.

\noindent{\textbf{Models need to localize and detect events accurately to perform well on \lcric}} To understand the importance of localizing and detecting events for \LVU models, we first train a TQN `event classifier' model to predict 1 of 12 events (Section~\ref{subsec:cricket_overview}) in a video clip. The model is trained on ground truth annotations from \lcricm and has a fairly high test accuracy of $84.79\%$. We then divide the clips into a contiguous sequence of `event segments' by either using ground truth segmentations from \lcricm (labeled `Ground Truth' in Table~\ref{tab:baselines:aggregation}) or by uniformly dividing the clips into 60 contiguous segments (labeled `Uniform' in Table~\ref{tab:baselines:aggregation}), as each 10-over clip contains 60 events. Finally, we leverage the learned `event classifier' model to generate event chains by sequentially predicting events on the `event segments' and evaluating different queries on these event chains. We report performance in Table~\ref{tab:baselines:aggregation} and make two observations -- \textbf{1)} Access to ground truth event segments leads to $\sim 20\%$ improvement over uniformly generated segments on binary queries and therefore shows the importance of event localization. \textbf{2)} While access to ground truth event segments leads to better performance, as it aids the event classifier in making more accurate predictions, the performance is still $\sim16\%$ worse than the human baseline on binary queries. Therefore, even with perfect event localization, models need performant event detection capabilities.

\noindent{\textbf{TQN performs better than MeMViT on all query types}} Table~\ref{tab:baselines} shows that TQN performs much better than MeMViT across different query types and training schemes ($+2\%$ on binary and multi-choice queries). While we expected MeMViT to perform better due to the clever caching mechanism in its architecture, we think adapting MeMViT with the TQN multi-query head led to the performance drop. We hope future work can better integrate the caching mechanism of MeMViT into the multi-query setting as it is infeasible to train one model for every new query.

\noindent{\textbf{Homogeneous training scheme leads to superior performance than Mixed training}} Our results in Table~\ref{tab:baselines} indicate that training different models for different query types leads to much better performance than training a single model for all queries. We also observe that performance on the regression query improves by $\sim$7 points and we posit that the model needs the extra representational capacity to answer the regression queries with high precision. 

\begin{table}[t!]
	\setlength{\tabcolsep}{3pt}

	\begin{center}
		\begin{tabular}{c c c c c ccc ccc}
			\toprule
            & \scriptsize \texttt{\#} &
        \multicolumn{1}{c}{\footnotesize\textbf{Model}} & \footnotesize\textbf{Training} & \footnotesize\makecell{\textbf{Binary}\\\textbf{Accuracy}} $\uparrow$ & \footnotesize\makecell{\textbf{Multiple}\\\textbf{Accuracy}} $\uparrow$ &  \footnotesize\makecell{\textbf{Regression}\\\textbf{L1 Norm}} $\uparrow$ \\
			\midrule
    & \scriptsize \texttt{1} &\footnotesize TQN & \footnotesize Mixed & \footnotesize \underline{57.68\%} & \footnotesize \underline{19.05\%} & \footnotesize 17.21 \\
    & \scriptsize \texttt{2} &\footnotesize MeMViT & \footnotesize Mixed & \footnotesize 54.31\% & \footnotesize 16.71\% & \footnotesize 21.79 \\
    & \scriptsize \texttt{3} &\footnotesize TQN & \footnotesize Homogeneous & \footnotesize \textbf{60.74\%} & \footnotesize \textbf{20.19\%} & \footnotesize \textbf{10.63} \\
    & \scriptsize \texttt{4} &\footnotesize MeMViT & \footnotesize Homogeneous & \footnotesize 56.53 \% & \footnotesize 17.79\% & \footnotesize \underline{11.95} \\
    
    \midrule
    & \scriptsize \texttt{5} &\footnotesize Human & & \footnotesize 96.34 \% & \footnotesize 96.29\% & \footnotesize 0.215 \\

            \bottomrule
		\end{tabular}
	\end{center}
	\vspace{-10pt}
	\caption{
    \textbf{Baseline performance on the full \lcric{} split with 10 over clips.}
    TQN outperforms MeMViT in different training schemes across different query types. We find that training models under the Homogeneous training scheme greatly improves performance, especially for the regression query.}
	\label{tab:baselines}
\end{table}

\begin{table}[t!]
	\setlength{\tabcolsep}{3pt}

	\begin{center}
		\begin{tabular}{c c c c c ccc ccc}
			\toprule
            & \scriptsize \texttt{\#} &
        \multicolumn{1}{c}{\footnotesize\textbf{Model}} & \footnotesize\makecell{\textbf{Clip}\\\textbf{Segments}} & \footnotesize\makecell{\textbf{Binary}\\\textbf{Accuracy}} $\uparrow$ & \footnotesize\makecell{\textbf{Multiple}\\\textbf{Accuracy}} $\uparrow$ &  \footnotesize\makecell{\textbf{Regression}\\\textbf{L1 Norm}} $\uparrow$ \\
			\midrule
    & \scriptsize \texttt{1} & \footnotesize TQN & \footnotesize Ground Truth & \footnotesize \textbf{80.34\%} & \footnotesize \textbf{36.32\%} & \footnotesize \textbf{6.89} \\
    & \scriptsize \texttt{2} & \footnotesize TQN & \footnotesize Uniform & \footnotesize 60.29\% & \footnotesize 15.18\% & \footnotesize 19.31 \\
    
    \midrule
    & \scriptsize \texttt{3} & \footnotesize Human &  & \footnotesize 96.34 \% & \footnotesize 96.29\% & \footnotesize 0.215 \\

            \bottomrule
		\end{tabular}
	\end{center}
	\vspace{-10pt}
	\caption{
    \textbf{Both localization and detection of events are important.} Using ground truth clip segments (aligned by ASAP) for event prediction leads to significant performance gain (Rows 1 vs 2). Improving event detection within clips can further improve performance (Rows 2 vs 3).
    }
	
	\label{tab:baselines:aggregation}

\end{table}

\section{Conclusion}

\noindent In this work, we introduce \ASAP, a fully automated annotation and video stream alignment pipeline for sports matches. \ASAP automatically aligns unlabeled videos of sports matches with corresponding dense annotations (\ie commentary) freely available on the web. We demonstrate the generality of \ASAP by aligning unlabeled matches of four very different sports with their corresponding annotations on the web. \ASAP is highly accurate (as judged by human annotators), and is robust to varying visual attributes, number of events, and length of plays. We then demonstrate \ASAP's potential to generate large-scale video datasets with \textit{no additional annotation cost} by generating \lcric, a large-scale long video understanding benchmark with over 1000 hours of densely annotated long Cricket videos (having an average sample length of $\sim$50 minutes). We extensively benchmark state-of-the-art \LVU models and establish a human baseline on \lcric. Our strong human baseline, coupled with the poor performance of state-of-the-art models, validates \lcric as an effective benchmark for the next generation of \LVU models. We hope that future work extends, improves, and leverages \ASAP to generate annotated video datasets at an unprecedented scale and cost efficiency.
\clearpage

{\small
\bibliographystyle{ieee_fullname}
\bibliography{main, misc_references}

\begin{thebibliography}{10}\itemsep=-1pt

\bibitem{posetrack}
Mykhaylo Andriluka, Umar Iqbal, Anton Milan, Eldar Insafutdinov, Leonid
  Pishchulin, Juergen Gall, and Bernt Schiele.
\newblock Posetrack: {A} benchmark for human pose estimation and tracking.
\newblock {\em CoRR}, abs/1710.10000, 2017.

\bibitem{bain2020condensed}
Max Bain, Arsha Nagrani, Andrew Brown, and Andrew Zisserman.
\newblock Condensed movies: Story based retrieval with contextual embeddings,
  2020.

\bibitem{cnn}
Y. Bengio and Yann Lecun.
\newblock Convolutional networks for images, speech, and time-series.
\newblock 11 1997.

\bibitem{timesformer}
Gedas Bertasius, Heng Wang, and Lorenzo Torresani.
\newblock Is space-time attention all you need for video understanding?
\newblock In {\em International Conference on Machine Learning}, pages
  813--824. PMLR, 2021.

\bibitem{byrne2008learning}
John~H. Byrne.
\newblock {\em Learning and Memory: A Comprehensive Reference}.
\newblock Elsevier, Oxford UK, 2008.

\bibitem{LoL}
Mohit~Bansal Cheng-Yang~Fu, Joon~Lee and Alexander~C. Berg.
\newblock Video highlight prediction using audience chat reactions.
\newblock In {\em EMNLP}, 2017.

\bibitem{meva}
Kellie Corona, Katie Osterdahl, Roderic Collins, and Anthony Hoogs.
\newblock Meva: A large-scale multiview, multimodal video dataset for activity
  detection.
\newblock In {\em Proceedings of the IEEE/CVF Winter Conference on Applications
  of Computer Vision (WACV)}, pages 1060--1068, January 2021.

\bibitem{amt}
Kevin Crowston.
\newblock Amazon mechanical turk: A research tool for organizations and
  information systems scholars.
\newblock In Anol Bhattacherjee and Brian Fitzgerald, editors, {\em Shaping the
  Future of ICT Research. Methods and Approaches}, pages 210--221, Berlin,
  Heidelberg, 2012. Springer Berlin Heidelberg.

\bibitem{redcaps}
Karan Desai, Gaurav Kaul, Zubin Aysola, and Justin Johnson.
\newblock {RedCaps: Web-curated image-text data created by the people, for the
  people}.
\newblock In {\em NeurIPS Datasets and Benchmarks}, 2021.

\bibitem{vit}
Alexey Dosovitskiy, Lucas Beyer, Alexander Kolesnikov, Dirk Weissenborn,
  Xiaohua Zhai, Thomas Unterthiner, Mostafa Dehghani, Matthias Minderer, Georg
  Heigold, Sylvain Gelly, et~al.
\newblock An image is worth 16x16 words: Transformers for image recognition at
  scale.
\newblock {\em arXiv preprint arXiv:2010.11929}, 2020.

\bibitem{ESPNfootball}
{ESPN}.
\newblock Espn soccer commentary, 2022.
\newblock \url{https://www.espn.in/football/commentary}.

\bibitem{ESPNCricinfo}
{ESPN}.
\newblock Espncricinfo, 2022.
\newblock \url{www.espncricinfo.com/}.

\bibitem{fan2021multiscale}
Haoqi Fan, Bo Xiong, Karttikeya Mangalam, Yanghao Li, Zhicheng Yan, Jitendra
  Malik, and Christoph Feichtenhofer.
\newblock Multiscale vision transformers.
\newblock In {\em Proceedings of the IEEE/CVF International Conference on
  Computer Vision}, pages 6824--6835, 2021.

\bibitem{x3d}
Christoph Feichtenhofer.
\newblock X3d: Expanding architectures for efficient video recognition.
\newblock In {\em Proc. {CVPR}}, 2020.

\bibitem{slowfast}
Christoph Feichtenhofer, Haoqi Fan, Jitendra Malik, and Kaiming He.
\newblock Slowfast networks for video recognition.
\newblock In {\em Proceedings of the IEEE/CVF international conference on
  computer vision}, pages 6202--6211, 2019.

\bibitem{gella2018dataset}
Spandana Gella, Mike Lewis, and Marcus Rohrbach.
\newblock A dataset for telling the stories of social media videos.
\newblock In {\em Proceedings of the 2018 Conference on Empirical Methods in
  Natural Language Processing}, pages 968--974, 2018.

\bibitem{OCR}
{Google}.
\newblock Google cloud optical character recognition, 2022.
\newblock \url{https://cloud.google.com/vision/docs/ocr}.

\bibitem{vidsrl}
Arka Gupta, Mark Yatskar, Ram Nevatia, and Aniruddha Kembhavi.
\newblock Visual semantic role labeling for video understanding.
\newblock In {\em CVPR 2021}, 2021.

\bibitem{MovieNet}
Qingqiu Huang, Yu Xiong, Anyi Rao, Jiaze Wang, and Dahua Lin.
\newblock Movienet: A holistic dataset for movie understanding.
\newblock In {\em Proceedings of the European Conference on Computer Vision
  (ECCV)}, 2020.

\bibitem{hudson2019gqa}
Drew~A Hudson and Christopher~D Manning.
\newblock Gqa: A new dataset for real-world visual reasoning and compositional
  question answering.
\newblock In {\em Proceedings of the IEEE/CVF conference on computer vision and
  pattern recognition}, pages 6700--6709, 2019.

\bibitem{tgif}
Yunseok Jang, Yale Song, Youngjae Yu, Youngjin Kim, and Gunhee Kim.
\newblock Tgif-qa: Toward spatio-temporal reasoning in visual question
  answering.
\newblock In {\em Proc. {CVPR}}, 2017.

\bibitem{johnson2017clevr}
Justin Johnson, Bharath Hariharan, Laurens Van Der~Maaten, Li Fei-Fei, C
  Lawrence~Zitnick, and Ross Girshick.
\newblock Clevr: A diagnostic dataset for compositional language and elementary
  visual reasoning.
\newblock In {\em Proceedings of the IEEE conference on computer vision and
  pattern recognition}, pages 2901--2910, 2017.

\bibitem{sports-1m}
Andrej Karpathy, George Toderici, Sanketh Shetty, Thomas Leung, Rahul
  Sukthankar, and Li Fei-Fei.
\newblock Large-scale video classification with convolutional neural networks.
\newblock In {\em 2014 IEEE Conference on Computer Vision and Pattern
  Recognition}, pages 1725--1732, 2014.

\bibitem{kazemi2012using}
Vahid Kazemi and Josephine Sullivan.
\newblock Using richer models for articulated pose estimation of footballers.
\newblock In {\em BMVC}, 2012.

\bibitem{krishna2017dense}
Ranjay Krishna, Kenji Hata, Frederic Ren, Li Fei-Fei, and Juan~Carlos Niebles.
\newblock Dense-captioning events in videos.
\newblock In {\em International Conference on Computer Vision (ICCV)}, 2017.

\bibitem{lei2018tvqa}
Jie Lei, Licheng Yu, Mohit Bansal, and Tamara~L Berg.
\newblock Tvqa: Localized, compositional video question answering.
\newblock In {\em EMNLP}, 2018.

\bibitem{lei2019tvqa}
Jie Lei, Licheng Yu, Tamara~L Berg, and Mohit Bansal.
\newblock Tvqa+: Spatio-temporal grounding for video question answering.
\newblock In {\em Tech Report, arXiv}, 2019.

\bibitem{ava_kinetics}
Ang Li, Meghana Thotakuri, David~A Ross, Jo{\~a}o Carreira, Alexander
  Vostrikov, and Andrew Zisserman.
\newblock The ava-kinetics localized human actions video dataset.
\newblock {\em arXiv preprint arXiv:2005.00214}, 2020.

\bibitem{liang2010}
Chao Liang, Yu Jiang, Jian Cheng, Changsheng Xu, Xiaowei Luo, Jinqiao Wang, Yu
  Fu, Hanqing Lu, and Jian Ma.
\newblock Personalized sports video customization for mobile devices.
\newblock In {\em Proceeding of International Conference on Multimedia Modeling
  (MMM)}, pages 614--625, 2010.

\bibitem{liang2010_context}
Chao Liang, Changsheng Xu, and Hanqing Lu.
\newblock Personalized sports video customization using content and context
  analysis.
\newblock In {\em International Journal of Digital Multimedia Broadcasting
  (IJDMB)}, 2010.

\bibitem{virat}
Sangmin Oh, Anthony Hoogs, Amitha Perera, Naresh Cuntoor, Chia-Chih Chen,
  Jong~Taek Lee, Saurajit Mukherjee, J.~K. Aggarwal, Hyungtae Lee, Larry Davis,
  Eran Swears, Xioyang Wang, Qiang Ji, Kishore Reddy, Mubarak Shah, Carl
  Vondrick, Hamed Pirsiavash, Deva Ramanan, Jenny Yuen, Antonio Torralba, Bi
  Song, Anesco Fong, Amit Roy-Chowdhury, and Mita Desai.
\newblock A large-scale benchmark dataset for event recognition in surveillance
  video.
\newblock In {\em CVPR 2011}, pages 3153--3160, 2011.

\bibitem{sbu_neurips2011}
Vicente Ordonez, Girish Kulkarni, and Tamara Berg.
\newblock Im2text: Describing images using 1 million captioned photographs.
\newblock In J. Shawe-Taylor, R. Zemel, P. Bartlett, F. Pereira, and K.~Q.
  Weinberger, editors, {\em Advances in Neural Information Processing Systems},
  volume~24. Curran Associates, Inc., 2011.

\bibitem{PontTuset_eccv2020}
Jordi Pont-Tuset, Jasper Uijlings, Soravit Changpinyo, Radu Soricut, and
  Vittorio Ferrari.
\newblock Connecting vision and language with localized narratives.
\newblock In {\em ECCV}, 2020.

\bibitem{sports-videos-in-the-wild-svw-a-video-dataset-for-sports-analysis}
Seyed~Morteza Safdarnejad, Xiaoming Liu, Lalita Udpa, Brooks Andrus, John Wood,
  and Dean Craven.
\newblock Sports videos in the wild (svw): A video dataset for sports analysis.
\newblock In {\em Proc. International Conference on Automatic Face and Gesture
  Recognition}, Ljubljana, Slovenia, May 2015.

\bibitem{conceptual_captions}
Piyush Sharma, Nan Ding, Sebastian Goodman, and Radu Soricut.
\newblock Conceptual captions: A cleaned, hypernymed, image alt-text dataset
  for automatic image captioning.
\newblock In {\em Proceedings of the 56th Annual Meeting of the Association for
  Computational Linguistics (Volume 1: Long Papers)}, pages 2556--2565,
  Melbourne, Australia, July 2018. Association for Computational Linguistics.

\bibitem{charades}
Gunnar~A Sigurdsson, G{\"u}l Varol, Xiaolong Wang, Ali Farhadi, Ivan Laptev,
  and Abhinav Gupta.
\newblock Hollywood in homes: Crowdsourcing data collection for activity
  understanding.
\newblock In {\em Proc. {ECCV}}, 2016.

\bibitem{ucf101}
Khurram Soomro, Amir~Roshan Zamir, and Mubarak Shah.
\newblock {UCF101:} {A} dataset of 101 human actions classes from videos in the
  wild.
\newblock {\em CoRR}, abs/1212.0402, 2012.

\bibitem{MovieQA}
Makarand Tapaswi, Yukun Zhu, Rainer Stiefelhagen, Antonio Torralba, Raquel
  Urtasun, and Sanja Fidler.
\newblock {MovieQA: Understanding Stories in Movies through
  Question-Answering}.
\newblock In {\em IEEE Conference on Computer Vision and Pattern Recognition
  (CVPR)}, 2016.

\bibitem{deepmind_football}
Karl Tuyls, Shayegan Omidshafiei, Paul Muller, Zhe Wang, Jerome Connor, Daniel
  Hennes, Ian Graham, William Spearman, Tim Waskett, Dafydd Steel, Pauline Luc,
  Adria Recasens, Alexandre Galashov, Gregory Thornton, Romuald Elie, Pablo
  Sprechmann, Pol Moreno, Kris Cao, Marta Garnelo, Praneet Dutta, Michal Valko,
  Nicolas Heess, Alex Bridgland, Julien Pérolat, Bart De~Vylder, S.~M.~Ali
  Eslami, Mark Rowland, Andrew Jaegle, Remi Munos, Trevor Back, Razia Ahamed,
  Simon Bouton, Nathalie Beauguerlange, Jackson Broshear, Thore Graepel, and
  Demis Hassabis.
\newblock Game plan: What ai can do for football, and what football can do for
  ai, May 2021.

\bibitem{vaswani2017attention}
Ashish Vaswani, Noam Shazeer, Niki Parmar, Jakob Uszkoreit, Llion Jones,
  Aidan~N Gomez, {\L}ukasz Kaiser, and Illia Polosukhin.
\newblock Attention is all you need.
\newblock In {\em Proc. {NeurIPS}}, 2017.

\bibitem{openttgames}
Roman Voeikov, Nikolay Falaleev, and Ruslan Baikulov.
\newblock Ttnet: Real-time temporal and spatial video analysis of table tennis.
\newblock {\em CoRR}, abs/2004.09927, 2020.

\bibitem{wu_longvideo}
Chao-Yuan Wu and Philipp Krahenbuhl.
\newblock Towards long-form video understanding.
\newblock In {\em Proceedings of the IEEE/CVF Conference on Computer Vision and
  Pattern Recognition}, pages 1884--1894, 2021.

\bibitem{memvit}
Chao-Yuan Wu, Yanghao Li, Karttikeya Mangalam, Haoqi Fan, Bo Xiong, Jitendra
  Malik, and Christoph Feichtenhofer.
\newblock Memvit: Memory-augmented multiscale vision transformer for efficient
  long-term video recognition.
\newblock {\em arXiv preprint arXiv:2201.08383}, 2022.

\bibitem{wu2022memvit}
Chao-Yuan Wu, Yanghao Li, Karttikeya Mangalam, Haoqi Fan, Bo Xiong, Jitendra
  Malik, and Christoph Feichtenhofer.
\newblock Memvit: Memory-augmented multiscale vision transformer for efficient
  long-term video recognition.
\newblock {\em arXiv preprint arXiv:2201.08383}, 2022.

\bibitem{xie2018rethinking}
Saining Xie, Chen Sun, Jonathan Huang, Zhuowen Tu, and Kevin Murphy.
\newblock Rethinking spatiotemporal feature learning: Speed-accuracy trade-offs
  in video classification.
\newblock In {\em Proc. {ECCV}}, 2018.

\bibitem{jwang2006}
C.S. Xu, J. Wang, K. Wan, Y. Li, and L. Duan.
\newblock Live sports event detection based on broadcast video and web-casting
  text.
\newblock In {\em Proceeding of ACM International Conference on Multimedia},
  page 221–230, 2006.

\bibitem{xu2016msr}
Jun Xu, Tao Mei, Ting Yao, and Yong Rui.
\newblock Msr-vtt: A large video description dataset for bridging video and
  language.
\newblock In {\em 2016 IEEE Conference on Computer Vision and Pattern
  Recognition (CVPR)}, pages 5288--5296, 2016.

\bibitem{zeng2016generation}
Kuo-Hao Zeng, Tseng-Hung Chen, Juan~Carlos Niebles, and Min Sun.
\newblock Title generation for user generated videos.
\newblock volume 9906, 10 2016.

\bibitem{zhang21}
Chuhan Zhang, Ankush Gupta, and Andrew Zisserman.
\newblock Temporal query networks for fine-grained video understanding.
\newblock {\em arXiv preprint}, 2021.

\bibitem{zhang2021temporal}
Chuhan Zhang, Ankush Gupta, and Andrew Zisserman.
\newblock Temporal query networks for fine-grained video understanding.
\newblock In {\em Proceedings of the IEEE/CVF Conference on Computer Vision and
  Pattern Recognition}, pages 4486--4496, 2021.

\bibitem{ZhXuCoCVPR18}
Luowei Zhou, Chenliang Xu, and Jason~J Corso.
\newblock Towards automatic learning of procedures from web instructional
  videos.
\newblock In {\em AAAI Conference on Artificial Intelligence}, 2018.

\end{thebibliography}
}

\clearpage

\appendix

\section{ASAP}
\label{appendix:asap}

\noindent{\textbf{Alignment Verification}}
We verify the accuracy of annotations made by \ASAP by providing human annotators with a clip containing a contiguous sequence of events, and asking them to provide the timestamps in the video for when each event occurred. Additionally, all scorecard information is masked in each provided clip.
\\

\noindent{\textbf{Verification of different sports}}
For cricket, we built an AMT interface and asked annotators to provide both the timestamps and events that occurred in a clip for over 1200 events to verify that both the \ASAP alignment process and video quality were sufficient, which we discuss more in Appendix \ref{appendix:lcric}. For verifying and demonstrating the generality of \ASAP pipeline, we annotate three different sports, namely, American football, football, and basketball, and verify it using a similar interface. Due to the limited mturk budget, we used two of the in-house annotators for the verification of these three sport's annotations by providing the humans with clips from 6 hours of match footage for each sport and had them verify (by annotating) 240 events for each sport. 
\\

\noindent{\textbf{American Football Alignment Issues}}
We note that the reason why the verification accuracy for American football in Figure \ref{fig:amt_plot} is lower than the other sports is because for most standard plays, the timestamps provided are for when the play started. However, when a team scores or is given a penalty, the timestamp provided for the next play is either the end of the play, or when it happened. We were only able to have \ASAP account for the touchdown instances, but not the penalty instances, which is generally what was marked incorrect during our verification process.
\\

\subsection{Qualitative Video}

We provide \textcolor{blue}{qualitative samples} under \texttt{'qualitative\_video\_asap'} directory in supplementary material by attaching some cricket match clips auto labelled using our ASAP pipeline.


\subsection{Annotation Event Details}
\label{appendix:events}
\noindent{\textbf{Events for Different Sports}}
In this section, we describe the events that we considered for each sport. 
\begin{itemize}
    \item \textbf{Cricket:} Each legal delivery was considered a valid event, where features such as the number of runs and the occurrence of a wide/out ball were marked as well. See Appendix \ref{appendix:lcric} for further details.
    \item \textbf{American Football:} Each play was considered a valid event, so we considered \textit{punts, field goals, complete passes, incomplete passes, run-plays, sacks, penalties,} and \textit{spikes} as distinct.
    \item \textbf{Football/Soccer:} There are no distinct, sequential plays in football, so we based our events off of online commentary. We mark \textit{shots off target, shots on target, shots on woodwork, goals, fouls, substitutions, yellow cards, red cards, corner kicks, free kicks, offsides, handballs,} and \textit{saved/blocked balls} as distinct events to be annotated and aligned.
    \item \textbf{Basketball:} Like football/soccer, there are no distinct plays that happen, so we mark \textit{fouls, jumper shots, layups, dunks, free throws}, and \textit{regular shots} as distinct events that we annotate and align. 
\end{itemize}

\noindent{\textbf{Granularity of Annotations}}
Because the aligned annotations for different sports rely on the timestamps provided by the online commentary source, we observe that different sports are annotated with varying levels of granularity. Thus, when we verify the accuracy of an aligned annotation, we account for these differing levels of granularity with different margins for error. For example, in football, annotations are provided at a minute-level, so if the human annotator marks the event as occurring anywhere outside that range, we consider the annotation to be incorrect; however, for sports like basketball, where annotation timestamps are given by the second, we provide a margin of error of $\pm 1$ second to the timestamp marked by the human. Similar to football, in cricket, an event lasts for 30-40 seconds, so if a human annotator is able to mark the event as occurring anywhere inside that range, we consider the annotation to be correct.


\subsection{Raw Videos Source}
All of the videos that we ran \ASAP through were found across YouTube. For cricket we used 131 videos, and for all other three sports we annotated 3 videos each. The average video length of a cricket match is 7.5 hrs while for the other sports it is 1.5 hrs. We also provide the links to all the videos annotated with the supplementary document.

\subsection{Examples of Frames filtered by ASAP}

\begin{figure}[t!]
    \centering
    \includegraphics[width=0.45\textwidth]{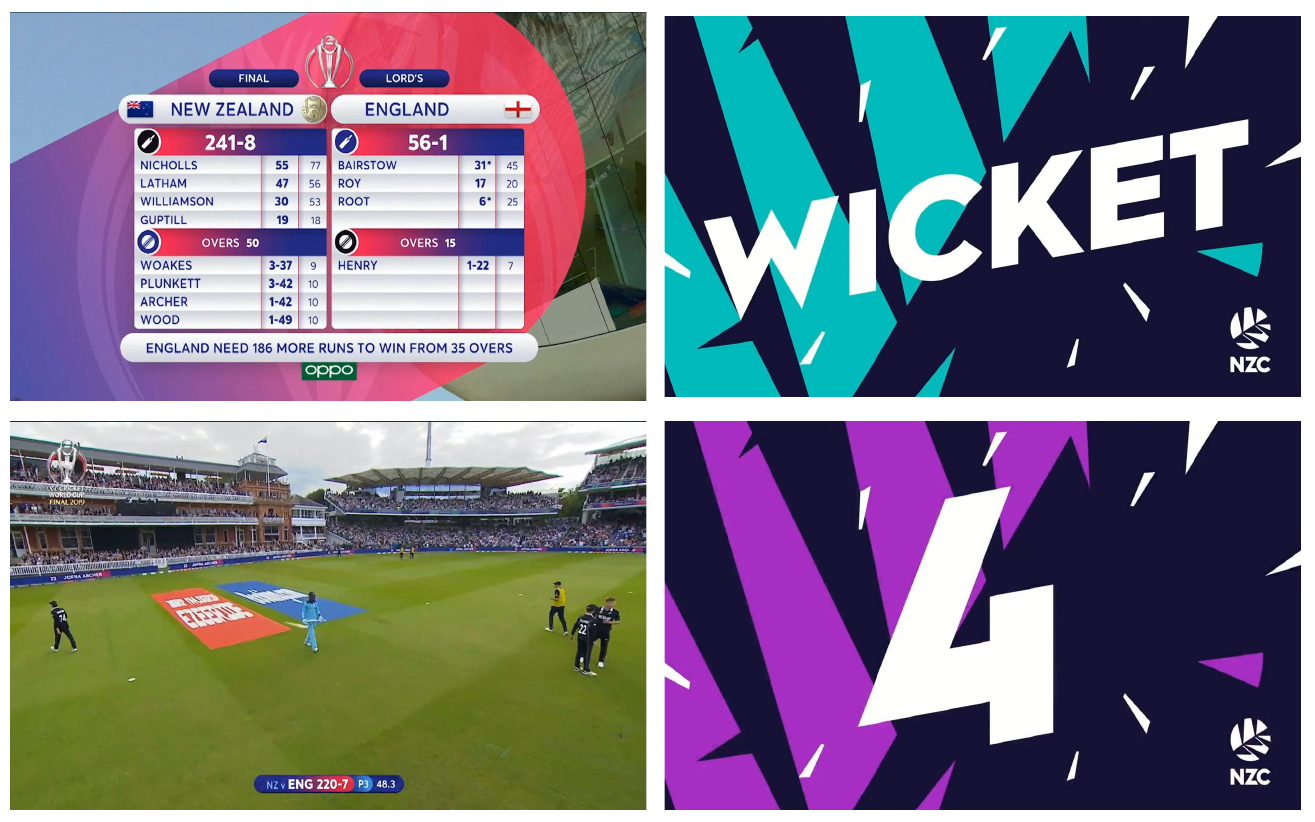}
    \captionof{figure}{Some examples of frames rejeced by ASAP. In all of these frames the scorecard information either get obstructed by screen overlays or shifted from their usual position.}
    \label{fig:wrong_frames}
\end{figure}

Some of the examples of frames being rejected by ASAP pipeline can be seen in Figure \ref{fig:wrong_frames}. As can be seen in most of these cases either the scorecard information is obstructed by some other text or some random data is present in its location.

\section{LCric}
\label{appendix:lcric}

\subsection{Primer on Cricket [\href{https://www.youtube.com/watch?v=g-beFHld19c}{Video}]}
\label{appendix:cric_primer}
In this section we further extend our primer to Cricket provided in Section~\ref{subsec:cricket_overview} by describing the Batting/Bowling phases, as well as the primary objective of the game. A brief overview video for the game explaining the game can be found in \href{https://www.youtube.com/watch?v=g-beFHld19c}{here}.
\\



\noindent{\textbf{Overview}}
Cricket is a ball-and-bat sport played by two teams of eleven players each. Cricket is scored by "runs", and at the end of the game, the team with the most scored "runs" wins. The game is played in an inning-format, where one team is batting, and the other team is fielding. We describe the two phases below.

\noindent{\textbf{Bowling Phase}}
When a team is in the bowling phase, all 11 players stay on the field. One of the players is designated as the bowler, and their job is to deliver the ball to the batter (hitter) on the batting team. If the ball is struck by the batsman, the remaining players, called fielders, try to prevent the ball from reaching the boundary of the field and return the ball back to the pitch area. A single over consists of six deliveries bowled by the same player, and each team delivers a set number of overs depending on the tournament type in their bowling phase. 
\\

\noindent{\textbf{Batting Phase}}
When is team is in the batting phase, only two players on the team stay on the field at a time. The batsman's job is to score runs and defend their wickets. A single run is scored when the batsman hits the ball and runs from one end of the pitch to another. Another way to score runs is to hit the ball to the boundary of the field, which is called the \textbf{'boundary'}, giving 4 or 6 runs to the batting team. In total, each batting team has 10 wickets.
\\

\noindent{\textbf{Objective}} 
During an inning, the batting team wants to score as many runs as possible, while the bowling team wants to take as many wickets as possible to stop the batting team from scoring. In most single-day matches, the bowling team will bowl for 50 overs before the teams switch roles for the second half of the game. At this point, the goal of the new batting team is to outscore the previous team in runs before 50 overs or before losing all of their wickets.
\\

\subsection{Training and Implementation Details}
We use consistent training schemes for both TQN~\cite{zhang21} and MeMViT~\cite{memvit} to provide a fair comparison between the two baselines. Both models were trained for 50 epochs on 4 V100 GPUs with a batch size of 4. We used a base learning rate of $LR = 0.01$ with the Adam optimizer and default hyperparameters.
\\

\noindent{\textbf{Baseline Implementations}}
For setting up TQN as a baseline, we used the official code provided by the authors with some minor modifications to the output heads for answering \lcric queries. For MeMViT, since there is no official implementation released at the time of writing, we implemented our own version using the same implementation details as the main paper. Our implementation is built on top of the official implementation of MViT~\cite{fan2021multiscale}, which is the base model used to create MeMViT.
\\

\noindent{\textbf{Attached Code}}
We have also attached the codebase with the supplementary zip and can be found in the 'code' repo. The codebase contains implementations for both our ASAP pipelines and our baseline implementations.
\\

\subsection{\lcric Queries}
\noindent{\textbf{Query Set Generation Algorithm}}
We describe our query set generation process in Algorithm \ref{alg:plan}, where we use logical operators and a set of possible atomic events form form different combinations of queries.
\\

\begin{algorithm}
\scriptsize
  \caption{\small \label{alg:plan} Query Set Generation}
  
  \label{alg:query_form}
  \var{\textcolor{gray}{\# The set of atomic events: [0,1,2,...,9,W,w]}} \\
  \textbf{Set of atomic events}: \texttt{$A_{e}$} 

  \var{\textcolor{gray}{\# The number of queries for the query set}} \\
  \textbf{Size of the query set}: \texttt{$n_q$} \\
        \texttt{query\_set} = [] \\
        \For{\textnormal{\texttt{i}} in \textnormal{\texttt{range($n_q$)}}} {
            \var{\textbf{\textcolor{blue}{\# Step A: getting raw operators and combinators choice}}} \\
            \texttt{num\_joins} $\sim$ [1,5]  \\
            \var{\textcolor{gray}{\# total length for operators set being sampled}} \\
            \var{\textcolor{gray} {for determining the query length}} 
            
            \texttt{ops} = \texttt{random.choices([\texttt{atleast()}, \texttt{atmost()}, \texttt{inrange()}], \texttt{num\_joins})} ~~ \var{\textcolor{gray}{\# sampling list of operators}} \\
            \texttt{combine\_op} = \texttt{random.choices([\texttt{and}, \texttt{or}], 1)} ~~ \var{\textcolor{gray}{\# sampling the combination operator}} \\

            \var{\textbf{\textcolor{blue}{\# Step B: instantiating a query for query set for matching }}}\\
            \var{\texttt{query} = []}\\
            \For{\textnormal{\texttt{op}} in \textnormal{\texttt{ops}} }{
                \var{\textcolor{gray}{\# specify lower bound for atleast/inrange ops}} \\
                \texttt{occ\_min} $\sim$ [1,10]  ~~ \\
                \var{\textcolor{gray}{\# specify upper bound bound for atmost/inrange ops}} \\
                \texttt{occ\_max} $\sim$ [\texttt{occ\_min},10] ~~ \\
                \var{\textcolor{gray}{\# sample atomic events in query}} \\
                \texttt{atomic\_event} $\sim$ [\texttt{$A_{e}$}] ~~  \\
                
                \var{\textcolor{gray}{\# Using the above variables for defining an occurrence pattern for atomic\_event}}
                
                \texttt{instanced\_op} = \texttt{op}(\texttt{occ\_min}, \texttt{occ\_max}, \texttt{atomic\_event})
                 \texttt{query.append(\texttt{instanced\_op})} \\
            }
            \texttt{final\_query} = \texttt{join\_op(query)} \\
            \texttt{query\_set.append(\texttt{final\_query})} \\
            
        }
\end{algorithm}

\noindent{\textbf{Binary Query Statistics}}
For our 10-over experiments, we formed a balanced set of 32 queries by taking queries from the set formed by Algorithm \ref{alg:query_form} and pruning them down so that given a random 10-over clip sampled uniformly from \lcric, there would be a $0.5 \pm 0.05$ probability that the query would hold true on that clip. We list the set of all such queries and their corresponding probabilities in Table 4.
\\

\begin{figure*} [t]
    \centering
    \includegraphics[width=0.75\linewidth]{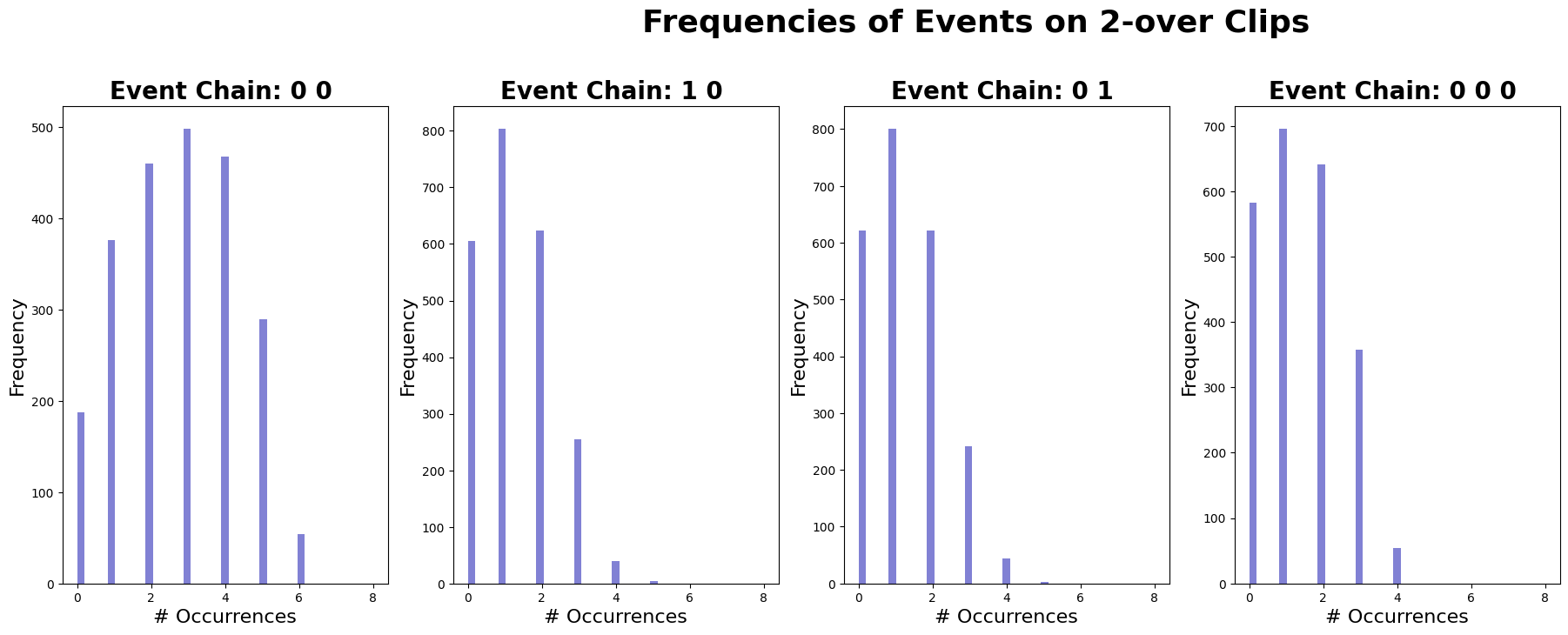}
    \includegraphics[width=0.75\linewidth]{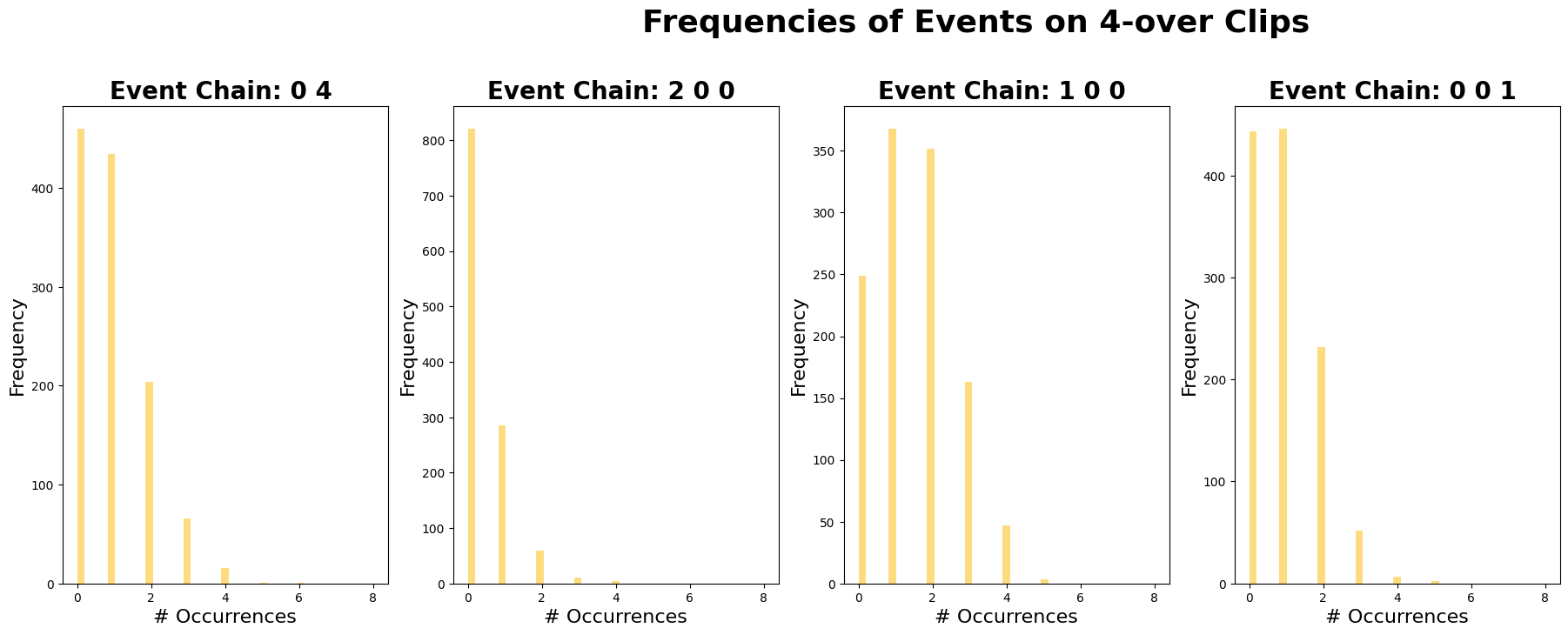}
    \includegraphics[width=0.75\linewidth]{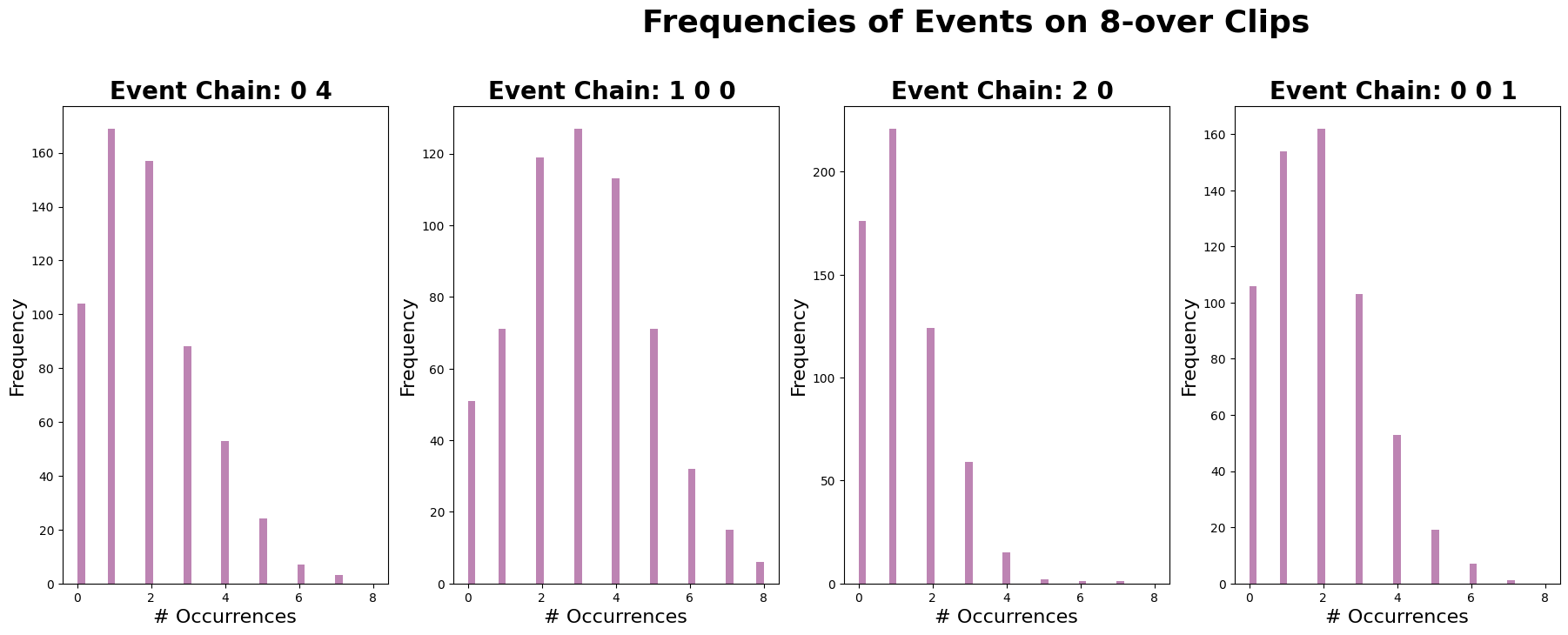}
    \includegraphics[width=0.75\linewidth]{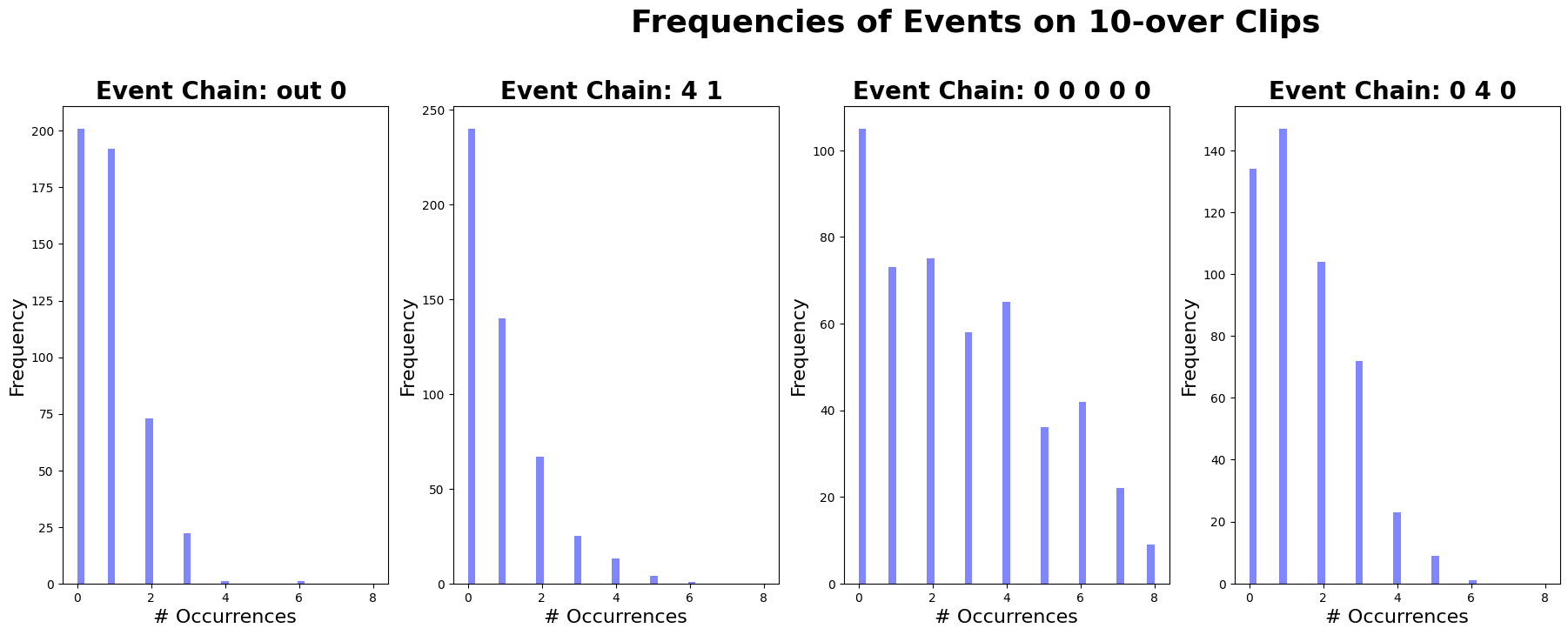}
    \captionof{figure}{Ground truth output frequencies to some of the queries used in multi-choice query type questions in the train set of \lcric for 2-over, 4-over, 8-over and 10-over clips.}
    \label{fig:multi_query_stats}
\end{figure*}

\noindent{\textbf{Multi-Choice Query Statistics}}
We also generated a set of multi-choice queries for our 10-over experiments. These queries include a mix of common and less common event chains that generally occur between 0-9 (inclusive) times within any 10-over clip. The frequency of occurrence of these clips within our train set is provided in Figure \ref{fig:multi_query_stats}.
\\

\subsection{AMT Interface}
\label{ssec:amt}
\begin{figure*}[t]
    \centering
    \includegraphics[width=0.9\linewidth]{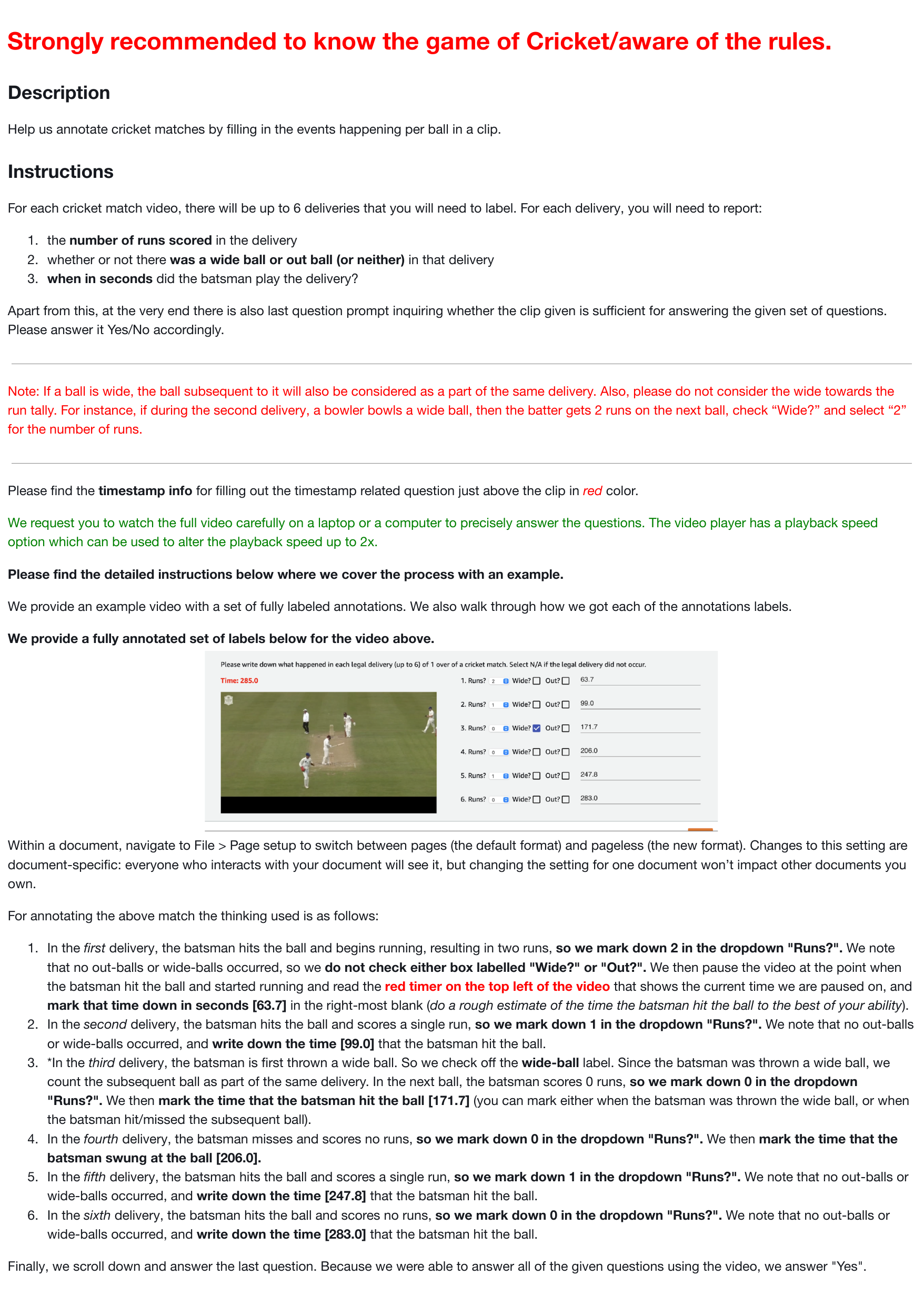}
    \caption{AMT instructions page given to annotators prior to starting the task.}
    \label{fig:amtinstructions}
\end{figure*}

We built an AMT interface for verifying \ASAP's alignment of cricket annotations to videos, with the full instructions and interface provided in Figure \ref{fig:amtinstructions}. 
\\

\begin{figure*}[t]
    \centering
    \includegraphics[width=0.9\linewidth]{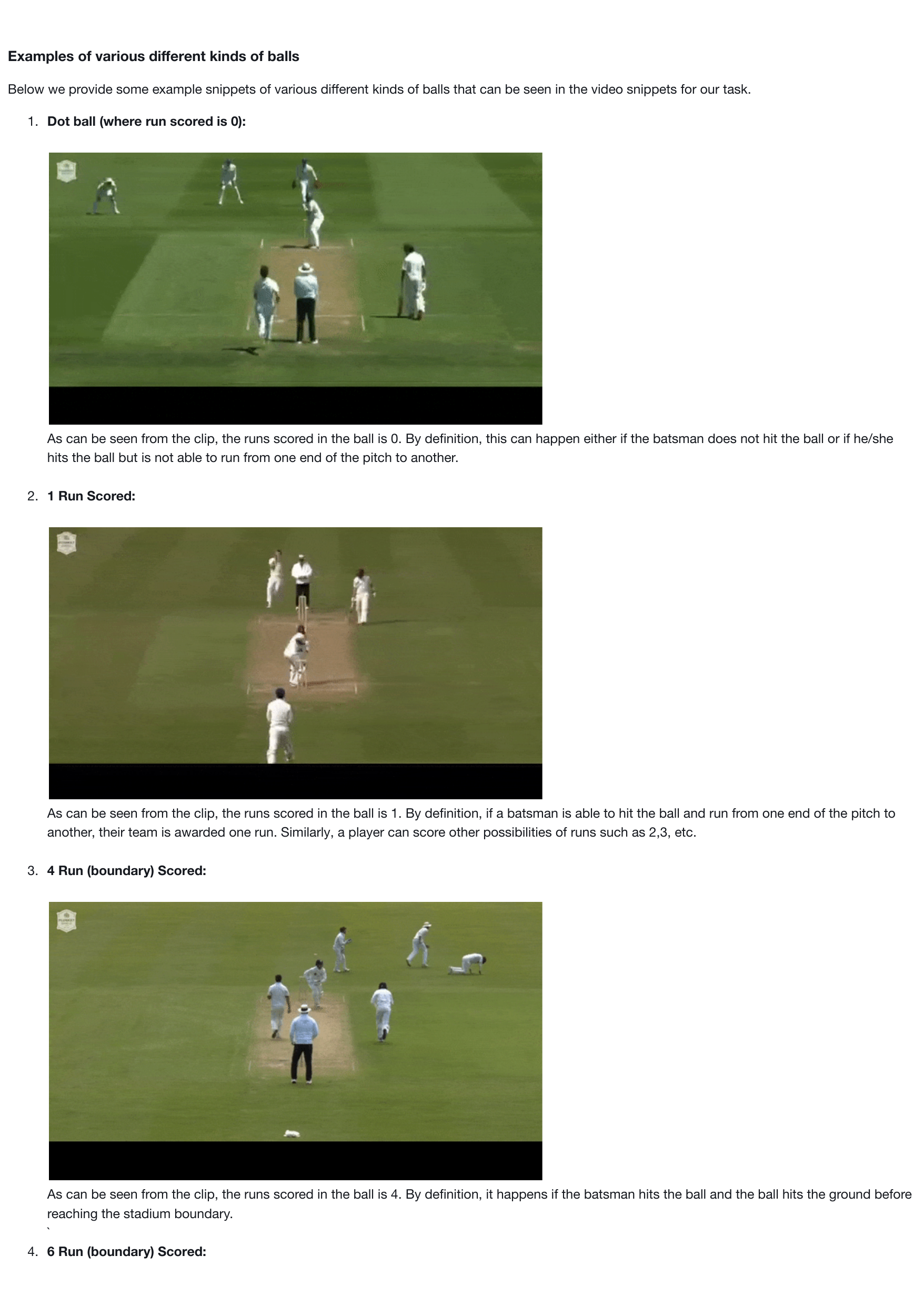}
    \caption{Instructions page for AMT interface for Cricket. Each of the 12 events is described in gif format.}
    \label{fig:amt_events1}
\end{figure*}

\begin{figure*}[t]
    \centering
    \includegraphics[width=0.9\linewidth]{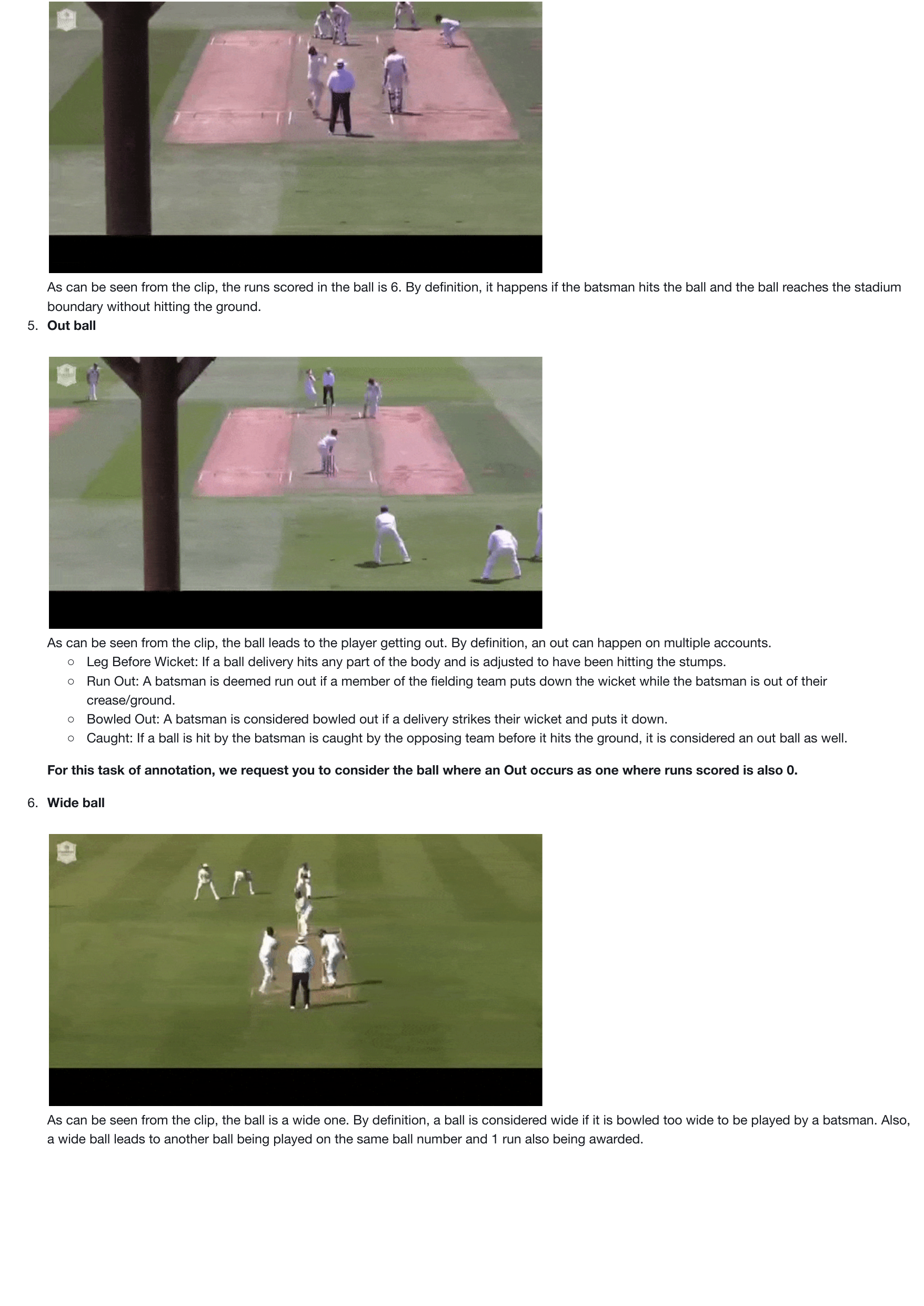}
    \caption{Instructions page for AMT interface for Cricket. Each of the 12 events is described in gif format.}
    \label{fig:amt_events2}
\end{figure*}

\noindent{\textbf{Instruction Details}}
Each annotator is given a set of instructions to read prior to beginning the main annotation task, called a HIT (Human Intelligence Task). For each task, the annotator is given a video clip from a sports match. The task is to classify each legal delivery/ball that occurred in the video, as well as the timestamp at which the annotator was able to gather enough information to answer this question. Additionally, we provide a set of examples for what each event looks like to the annotators, as well as a fully annotated example and video, as shown in Figure \ref{fig:amt_events1}, \ref{fig:amt_events2}.
\\

\noindent{\textbf{Task Interface Details}}
Each HIT contains a 1-over video and 6 rows, each corresponding to a legal delivery that occurred in the video. Each row consists of a dropdown for inputting the number of runs scored in that delivery, a checkbox for indicating an out ball occurred, a checkbox for indicating a wide ball occurred, and a field for writing the timestamp at which this information can be found. Figure \ref{fig:amtinstructions} shows what the annotators initially see, as well as an example of how to fill it out. 
\\

\noindent{\textbf{\lcric Annotation Verification}}
A total of 205 overs with 1230 events spanning $\sim$1000 minutes were labeled by human annotators and compared to ground truth annotations from ESPNCricinfo. For each ball, we consider an event annotation to be correct if it was classified completely correctly. The timestamp annotation is marked as correct if it occurred anytime within the timestamp range specified by the ground truth $\pm 1$ seconds.
\\

\noindent{\textbf{\lcric Annotation Statistics}}
We found that in total, $1185/1230 (96.34\%)$ of balls were classified correctly, while $1213/1230 (98.62\%)$ of ball timestamps were marked correctly. Additionally, assuming human annotators can aggregate and reason easily with logic, we aggregate their annotations to answer queries in our test set, which provides our human baseline. We find that the human annotations achieve an accuracy of $5541/5740 (96.53\%)$ on the test query set -- exceeding the TQN and MemViT baselines by a large margin. 
\begin{table}[ht]
    \scriptsize
    \begin{center}
		\begin{tabular}{| p{6.0cm} | p{2.0cm} |}
			\hline
            \textbf{Queries} & \textbf{GT probability}
            \\
            \hline
            atmost 7 1's & 0.451
            \\
            \hline
            atleast 4 4's & 0.523
            \\
            \hline
            atleast 5 1's AND atleast 3 4's & 0.528
            \\
            \hline
            atleast 2 2's AND atleast 3 4's & 0.452
            \\
            \hline
            atleast 4 4's AND atmost 5 o's & 0.452
            \\
            \hline
            atleast 4 4's AND atmost 3 5's & 0.456
            \\
            \hline
            atleast 4 2's OR atmost 2 4's & 0.539
            \\
            \hline
            atleast 4 3's OR atmost 3 4's & 0.544
            \\
            \hline
            atleast 5 2's OR atleast 4 4's & 0.526
            \\
            \hline
            atleast 3 2's OR atleast 2 w's & 0.485
            \\
            \hline
            atmost 3 4's AND atmost 2 6's & 0.529
            \\
            \hline
            atmost 3 4's AND atmost 3 7's & 0.544
            \\
            \hline
            atmost 2 0's OR atmost 3 4's & 0.544
            \\
            \hline
            2 inrange [1, 6] AND 4 inrange [1, 4] & 0.539
            \\
            \hline
            4 inrange [1, 6] AND o inrange [1, 4] & 0.555
            \\
            \hline
            1 inrange [2, 7] OR 2 inrange [4, 5] & 0.506
            \\
            \hline
            1 inrange [1, 2] OR 2 inrange [2, 3] & 0.458
            \\
            \hline
            atleast 2 1's AND atleast 2 2's AND atleast 2 4's & 0.542
            \\
            \hline
            atleast 4 4's OR atleast 4 o's OR atleast 4 w's & 0.493
            \\
            \hline
            atleast 5 2's OR atleast 4 4's OR atleast 3 6's & 0.535
            \\
            \hline
            atmost 4 3's AND atmost 3 4's AND atmost 2 5's & 0.544
            \\
            \hline
            atmost 4 2's AND atleast 3 4's AND atmost 4 w's & 0.546
            \\
            \hline
            atmost 5 1's OR atleast 5 3's OR atmost 2 4's & 0.504
            \\
            \hline
            atmost 3 0's OR atleast 5 3's OR atmost 3 4's & 0.544
            \\
            \hline
            atmost 3 0's OR atmost 4 1's OR atmost 2 4's & 0.472
            \\
            \hline
            atmost 2 0's OR atmost 5 1's OR atmost 2 4's & 0.504
            \\
            \hline
            1 inrange [2, 6] OR 2 inrange [3, 4] OR 3 inrange [6, 7] & 0.528
            \\
            \hline
            atleast 4 0's AND atleast 3 1's AND atleast 2 2's AND atleast 2 4's & 0.52
            \\
            \hline
            atleast 4 4's OR atleast 2 5's OR atleast 2 6's OR atleast 4 o's & 0.518
            \\
            \hline
            atmost 3 2's AND atmost 4 4's AND atmost 3 6's AND atmost 5 w's & 0.539
            \\
            \hline
            6 inrange [1, 7] OR 8 inrange [2, 4] OR o inrange [2, 3] OR w inrange [6, 7] & 0.494
            \\
            \hline
            1 inrange [1, 6] OR 5 inrange [1, 2] OR o inrange [3, 6] OR w inrange [4, 6] & 0.511
            \\
            \bottomrule
		\end{tabular}
		\caption{\small{The binary choice query set used for 10 over experiments and their associated ground truth (GT) probability of occurrence in the \lcric train set. }}
	\end{center}
	\label{tab:all_queries}
\end{table}


\end{document}